Decentralized Deep Reinforcement Learning for Network Level Traffic Signal Control

By

JIN GUO
THESIS

Submitted in partial satisfaction of the requirements for the degree of

MASTER OF SCIENCE

in

Transportation Technology and Policy

in the

OFFICE OF GRADUATE STUDIES

of the

UNIVERSITY OF CALIFORNIA

DAVIS

Approved:

\_\_\_\_\_\_\_\_\_\_\_\_\_\_\_\_\_\_\_\_\_\_\_\_\_\_\_\_\_\_\_\_\_\_\_\_\_\_\_\_
Prof. Michael Zhang, Chair

\_\_\_\_\_\_\_\_\_\_\_\_\_\_\_\_\_\_\_\_\_\_\_\_\_\_\_\_\_\_\_\_\_\_\_\_\_\_\_\_
Prof. Yueyue Fan

\_\_\_\_\_\_\_\_\_\_\_\_\_\_\_\_\_\_\_\_\_\_\_\_\_\_\_\_\_\_\_\_\_\_\_\_\_\_\_\_
Prof. Dipak Ghosal

Committee in Charge

2019

*To my father*

*You are always with me*

# CONTENTS

# LIST OF FIGURES

# LIST OF TABLES

# ACKNOWLEDGMENT

My deepest gratitude comes to my advisor, Prof. Michael Zhang, for his academic guidance and financial support. I am extremely fortunate to be advised by Prof. Zhang throughout my graduate research. His constant advice guides me to the wonderful world of transportation research, making me recognize the rigorousness and persistence of academic research.

Thanks especially also to Prof. Yueyue Fan for serving as my co-advisor and the committee member of my thesis. Thank you so much for encouraging me to study at UC Davis, your support made this thesis possible. I am also grateful to Prof. Dipak Ghosal, who provided valuable suggestions during the C3PO meeting and my thesis writing.

I would like to express my gratitude to the Institute of Transportation Studies at UC Davis and the National Center for Sustainable Transportation, for offering the fellowship and much more support to me. Thanks to the great faculties, researchers, peer students and the most warm-hearted coordinator, Annemarie Schaaf at ITS, UC Davis.

Last, I dedicate this thesis to my family and my beloved fiancée Shu Liu. You always stand with me no matter ups and downs, laughs and tears.

# ABSTRACT

Taming traffic congestion and its negative social-environmental impacts has long been a daunting problem for traffic engineers and transportation planners. Besides adding road capacity, advanced traffic management, in this context, more intelligent signal control is a promising tool to reduce travel delay, fuel use and air pollution. In this thesis, I propose a family of fully decentralized deep multi-agent reinforcement learning (MARL) algorithms to achieve high, real-time performance in network-level traffic signal control. In this approach, each intersection is modeled as an agent that plays a Markovian Game against the other intersection nodes in a traffic signal network modeled as an undirected graph, to approach the optimal reduction in delay. Following Partially Observable Markov Decision Processes (POMDPs), there are 3 levels of communication schemes between adjacent learning agents: independent deep Q-leaning (IDQL), shared states reinforcement learning (S2RL) and a shared states & rewards version of S2RL--S2R2L. In these 3 variants of decentralized MARL schemes, individual agent trains its local deep Q network (DQN) separately, enhanced by convergence-guaranteed techniques like double DQN, prioritized experience replay, multi-step bootstrapping, etc.

To test the performance of the proposed three MARL algorithms, a SUMO-based simulation platform is developed to mimic the traffic evolution of the real world. Fed with random traffic demand between permitted OD pairs, a 4×4 Manhattan-style grid network is set up as the testbed, two different vehicle arrival rates are generated for model training and testing. The experiment results show that S2R2L has a quicker convergence rate and better convergent performance than IDQL and S2RL in the training process. Moreover, three MARL schemes all reveal exceptional generalization abilities. Their testing results surpass the benchmark Max Pressure (MP) algorithm, under the criteria of average vehicle delay, network-level queue length and fuel consumption rate. Notably, S2R2L has the best testing performance of reducing 34.55% traffic delay and dissipating 10.91% queue length compared with MP.

Furthermore, to close the "reality gap" in RL research, I propose a policy-based RL scheme. The Importance Sampling technique updates the parameters of the real-world policy network using SARSA trajectories sampled from simulation. The full implementation of this is left to future work.

# 1. Introduction

## 1.1 Motivation

In 2018, the total cost of traffic congestion in the US was $87 billion, each commuter lost $1,348 on average due to the extra time spent behind wheels (CNBC, 2019). By the end of 2018, the number of registered vehicles in the US increased to 276.1 million from the 2014 statistic of 248.9 million (US DOT, 2019), with an increase of 7.7% (2.97 to 3.21 trillion miles) in total Vehicle Miles Traveled (VMT) (FHWA, 2019). Building more roads and increasing their capacities were historically used as the main approaches to address traffic congestion and its negative externalities. However, due to cost and environmental concerns, the total mileage of public roads only increased slightly by 1.8 % (FHWA, 2018) in the last five years, far behind the increase in travel demand.

More effective organization and management of traffic flow is a promising way to alleviate congestion, making better use of limited road capacity. Compared with freeway traffic flow, whose congestion formation and evolution is easier to be modeled (e,g, Zhang, 2002), urban traffic is more complicated and more stochastic with mixed autonomy (Aditya Teja, Viswanath, & Krishna, 2008), recurring (e.g., high demand in peak hours) and non-recurring disturbances (e.g. accidents and road construction) that runs the risk of uncontrollability (Jeihani, James, Saka, & Ardeshiri, 2015). Traffic signal control (TSC) is the core of urban traffic control, it functions in assigning road rights to vehicles, pedestrians, bicyclists and other road users from different directions. A well-designed traffic signal promises to effectively maximize the traffic throughput at the intersection (El-Tantawy, Abdulhai, & Abdelgawad, 2013), to reduce the frequency and severity of certain types of collisions (Sunkari, 2004), and to provide safe accessibility for vulnerable road users (Sarkar, Sahoo, & Sahoo, 2012).

There are even more benefits we can expect from optimized signal timing, not only for the mobility system but also for other dynamic systems coupled with it. As a physical backbone to civilization, transportation accounts for 28% of total energy consumption (US EIA, 2019), and 29 % of total Green House Gas (GHG) emission in the US (EPA, 2019), which is the largest portion among all the sectors. The National Transportation Operations Coalition (NTOC) reports that if the overall quality of traffic signal operations could be improved from the current 'D level' to 'A level', the US mobility system would see a decrease in traffic delay ranging from 15% to 45% and commuters will save up to 25% travel time. Up-to-date signal timing is slated to cut down motor fuels by 10%, nationwide, this would amount to a savings of almost 170 billion gallons per year. Consequently, up to 22% of harmful emissions would be reduced.

Three revolutions (shared, electric, automated) (Sperling, 2018) happening in transportation calls for more efficient, safer and less polluting infrastructure utilization. To accommodate increasing mobility demands with limited road capacity and to prevent the corresponding infrastructure degradation, significant efforts have been persistently and dedicatedly donated from researchers. Rapid development in the areas of communications and computing have full the gears for advanced traffic signal control, a nutshell of research progress will be reviewed in the next part.

## 1.2 Overview of Urban Traffic Signal Control Strategies

Control strategies are the backbone of TSC, supporting the system to operate efficiently and robustly. Promoted by the rapid development in the areas of telecommunication and computing, remarkable progress has been made in designing more advanced algorithms. Thus, this part provides an overview of how TSC strategies heading towards better intelligence.

### 1.2.1 Terminology and Key Definitions

*Actuated Signal Control* – Categorized as fully-actuated and semi-actuated, depending on whether the flexible timing mechanism is entirely or partially triggered by real-time traffic.

*Arrival rate* – The mean of a statistical distribution of how many vehicles arrive in a given time interval.

*Cycle length* – The total time to complete one sequence of signal indications for all movements at an intersection.

*Delay* – The additional time travelers (drivers, bicyclists, pedestrians, etc) lose at an intersection (or during all his trip) due to circumstances that impede the desirable movement of traffic (Source: AASHTO Glossary).

*Fixed-Time (Pre-timed) Signal Control* – A predetermined timing is assigned to each movement regardless of the change of any other conditions (e.g., traffic demand, weather, time of day).

*Flow rate* – The volume of vehicles passing a road segment or a single point in a given time interval.

*Flow ratio* – The ratio of the actual flow rate to the saturation flow rate.

*Isolated intersection* - An intersection located outside the influence of and not coordinated with other signalized intersections, commonly one mile or more from other signalized intersections. (Source: Caltrans Traffic Signal Operations Manual)

*Lost time* – The time period during which there is no vehicle movement even the signal shows green.

*Maximum/Minimum Green* – The allowed maximum/minimum time of a green phase.

*Offset* – The time difference between the begin of green of a phase and a reference point.

*Permitted turn* – Left or right turn at a signalized intersection that is made against an opposing or conflicting vehicular or pedestrian flow.

*Phase* –right-of-way assigned to an independent traffic movement.

*Queue* – A line of road users (typically vehicles) waiting to be served at an intersection.

***Split*** – The time assigned to a stage/phase during coordinated control.

***Stage*** – A time period during which no-conflicting traffic movements share the right of way at an intersection, typically one signal cycle consists of several stages and each stage consists of several phases.

**1.2.2 Review of Classical Signal Control Strategies**

Classical control strategies refer to what has been widely used in real-world scenarios, proven to handle both under-saturated and saturated traffic flow effectively and safely. These methods mainly rely on prespecified models (fixed-time, actuated control) or dynamic programming (adaptive control), where mathematical reasoning and control stability are rigorously obeyed. Topologically, a traffic signal network consists of signal groups at individual intersections and the linkages between them. Following the point-line-plane evolution, three levels of control strategies, i.e., isolated intersection, arterial road, traffic network, will be reviewed accordingly.

**Isolated Intersection**

*Fixed Timing strategies* are those derived offline using historical traffic statistics (arrival rates, turn ratios, etc) to give a uniform or time-of-day based timing schedule. They are ideally suited to closely spaced intersections where traffic demand and patterns are consistent during a long time period. An original calculation method for cycle length and green time distribution was proposed by Webster (Webster, 1958). There are mainly two types of fixed timing: *Stage-based strategies* predetermine combinations of phases before calculating the optimal cycle length and splits in order to minimize estimated delay or queue length, SIGSET(ALLSOP, 1971) and SIGCAP(Allsop, 1976) are two principal schemes of them. *Phase based strategies* not only optimize cycle length and splits, but also recommend the best stage design (combinations of phases), which without doubt increases the computational complexity, leading to a binary-mixed-integer-linear-programming problem, specific branch-and-bound solutions can be found in (Improta & Cantarella, 1984).

*Fixed Timing Strategies* have several advantages, e.g., this detection-free method requires no sensors, thus has the lowest cost to be installed and maintained. Also, the stability of fixed timing control is well proven in under-saturated flow conditions (Yin, 2008), making the system robust and controllable. Finally, for the purpose of coordination with adjacent signals, fixed timing provides enough resilience for offset, since the starting and ending of green indications are known. While pre-timed control has an innate defect that it cannot deal with unexpected fluctuations in traffic flows, if vehicles' arrival rates have a different distribution with historical surveys, the level of service (LOS) of the intersection tends to degrade even fail down.

*Actuated Control Strategies* use sensors (e.g., loop detectors, radars, cameras) to detect vehicle presence and movements, then assign green indications to corresponding phases. There are two main types of actuated control strategies, i.e., *semi-actuated* and *fully-actuated*. *Semi-actuated* control distributes detectors only on some approaches to an intersection, most commonly, branches with low-demand traffic. The phases associated with the major-road through movements are operated as "non-actuated", which means these movements are always permitted unless the controller receives calls from conflicting phases. That is to say, traffic on minor roads is served after they are detected and the elapsed time of the current phase reaches planned minimum green.

*Semi-actuated* control is particularly suitable for an intersection connecting an arterial road and a minor road, because compared with *fixed timing strategies*, it does reduce the delay of dominant through traffic. However, if the demands of two conflicting phases are comparable and traffic volumes have a large variance during the day, *semi-actuated* control tends to block the through traffic on the major road by just giving the minimum green time since the minor roads send requests frequently.

To fix the bias problem caused by *semi-actuated* control, *fully-actuated* strategies put detectors under each approach, to assign road right in a fairer way. The concept of the signal cycle becomes ambiguous in *fully-actuated* control since some phases can be extended or skipped depending on traffic volumes. First, a minimum green time is assigned to a phase or several non-conflicting phases, if no vehicle passes the pulse or presence detector during the predefined passage time (typically 2.0 to 4.5 seconds, depending on detection zone length and location), these phases will be "gap out", arbitrarily switched to phases on demand. Maximum green time is also set to prevent one phase with continuous high volume from being always activated, causing traffic in other directions to get stuck for long time. Once the timer touches the ceiling value, corresponding phases get "max out". More practical settings are supposed to be considered to serve different volume-density features, gap reduction and variable initial green are used to adjust previously fixed passage time and minimum green correspondingly (Akcelik, 1994). For more detailed designs, I refer to FHWA Traffic Signal Timing Manual, Chapter 5 (Fhwa, 2008) for detailed reference.

**Arterial Road**

*Coordinated control* is the main idea for traffic signals along an arterial road collaborating to provide smoother traffic, which, reduces travel times, fuel consumptions and air pollution. Practically, there are several warrants identifying the necessity of coordination, e.g., 1) Traffic signals are within 0.5 miles of each other along an arterial. 2) The arriving traffic includes platoons formed by the release of vehicles from the upstream intersection. Three fundamental parameters shape a coordinated signal system: cycle length, splits, offset. Cycle lengths of coordinated signals are required to be identical, with exceptions of "half-length" or "double-length", therefore the beginning of coordinated phases could synchronize with the background cycle maintained by a master clock within two cycles. Offsets define the difference of coordinated phases' starting time compared with a reference point, which is commonly the activation time of the corresponding phase of a head or rear intersection. Once a cycle length and offsets are determined, green times (the difference between the cycle length and the lost time) of phases are divided. Typically, splits for coordinated phases are guaranteed with a minimum guarantee, the rest of the time is then allocated to accommodate calls of other movements, before "forced off" to serve the coordinated one. Depending on whether there exists skips of non-coordinated phases and floating splits, coordinated strategies compose of *fixed-coordinated* ones and *actuated-activated* ones.

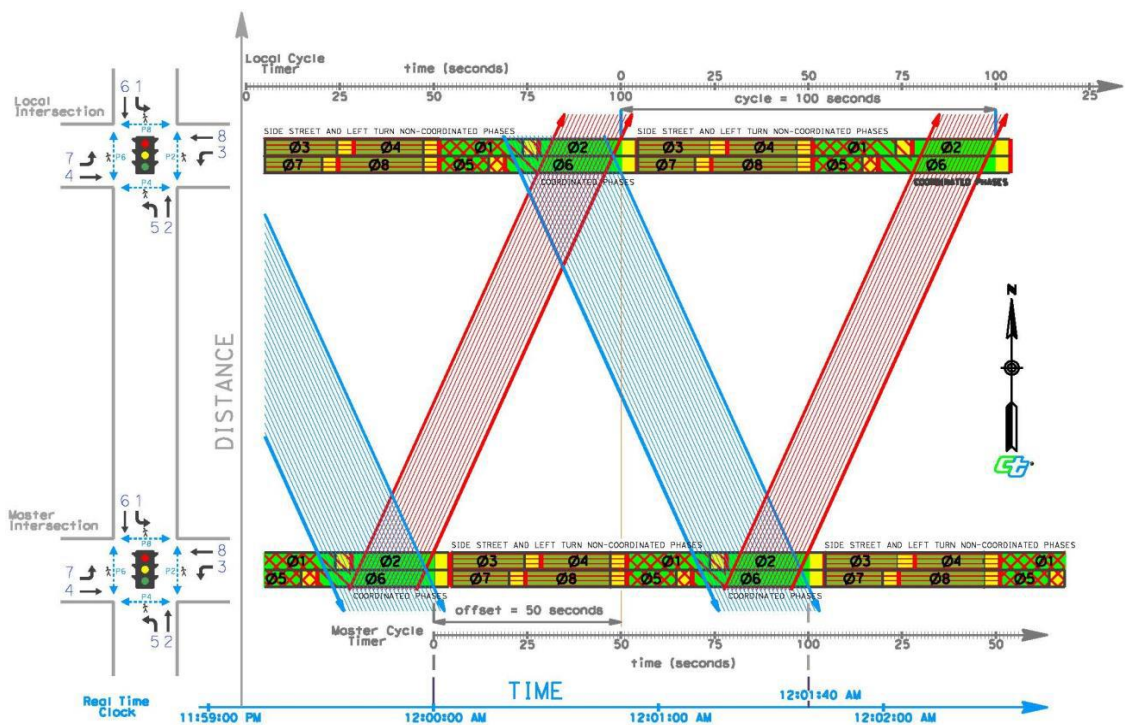

**Figure 1: Time-Space Diagram of a Coordinated Two-Way Street**
*(Reference: Caltrans Traffic Signals Operation Manual 2017)*

MAXBAND is a representative of *fixed-coordinated* strategies with the first version of (Little, 1966) and a MULTIBAND version proposed in (Stamatiadis & Gartner, 1997). As shown in **Fig. 1**, phase 2 (northbound through) and phase 6 (southbound through) are coordinated phases of these two adjacent intersections. MAXBAND defines the offsets of the upper signal, to create and maximize the bandwidth of a two-way "green wave" (marked as blue and red sloping bars), during which vehicles can travel without stopping in a recommended speed range. With the existence of protected left turn, the inbound and outbound bandwidths $\bar{b}$ and $b$ are calculated using a binary-mixed-integer-linear programming.

*Fixed-coordinated* strategies have the same drawbacks as fixed isolated intersection timing that cannot respond well to fluctuated flows, hence *actuated-coordinated* schemes are introduced to achieve better performance. Rather than operated in actuated mode individually, signals along an arterial assign their green time to non-coordinated phases according to their requests while maintaining an allotted minimum interval for coordinated phases. This requires a dedicated broadband wireless infrastructure or hardwired interconnect cables to exchange real-time information through signals, and fully equipped detectors to apply the actuated mode.

**Network Level**

Rather than just give priority to arterial traffic, network-level TSC extends the principles of adaptation and coordination to all the connected intersections (nodes) in a road network. These strategies can be classified into *centralized* and *decentralized* (or distributed), depending on whether a central control unit is used.

The calculation and online optimization of signal timing in traffic-response systems like SCOOT (Robertson & Bretherton, 1991), OPAC (Gartner, Pooran, & Andrews, 2001), RHODES (Mirchandani & Head, 2001), etc are operated in a distributed way, each intersection uses upstream flow measurements (OPAC), a queue model (SCOOT), or a dynamic network model (RHODES) to estimate link flows over a rolling horizon, thus to tune the timing schedule to minimize the total delay and stops.

The other philosophy is centralized control, where local detectors send real-time traffic patterns to a central controller, then the control unit recommends a network-level coordination schedule based on a globally optimized solution. Well-known centralized systems are ALLONS-D (Porche & Lafortune, 1999) and TUC (Aboudolas, Papageorgiou, Kouvelas, & Kosmatopoulos, 2010). Compared with distributed methods, centralized ones definitely require robust communication infrastructure and more computational resources that would increase the deployment cost.

**1.2.3 Review of Intelligent Algorithms in Network Level Signal Control**

I define intelligent control algorithms as what depend mainly on computational intelligence, generating adaptive control strategies based on trial-and-error feedback. There are quite many variants of artificial intelligence (AI) based algorithms for advanced TSC, since the background area is broad and has been the hottest research topic for more than two decades. Therefore, I just give a quite brief review on TSC strategies coupled with the three mostly applied paradigms, i.e., Fuzzy Logic (FL), Genetic Algorithm (GA) and Reinforcement Learning (RL), and scenarios of network-level control are given priority, more details concerning RL-based algorithms will be introduced in Chapter 4.

**Fuzzy Logic (FL)**, belonging to the category of Soft Computing, was first proposed by Zadeh (1965) to tackle complex control tasks based on "degree of truth" rather than Boolean logic. Pappis & Mamdani (1977) are the pioneers to transfer fuzzy rules to TSC, the signals of a two-direction one-way intersection are scheduled by a controller with three inputs, whether to extend the green time of the current phase is determined by vague reasoning, which compares the extensions with the highest degree of confidence. Nakatsuyama et al (1985) extended the application scenario to two adjacent intersections at an arterial road, where the fixed duration of green extension is determined by upstream traffic patterns. With more complicated intersection configurations, variable phase length and sequences are available in the multiple intersection setting discussed in (Lee & Lee-Kwang, 1999), where each junction makes decisions according to its own and neighboring traffic situations. To deal with different distributions of flow rates, researchers made attempts to design more sophisticated FL controllers, i.e., the two-layer FL architecture proposed by (Zhang, et al, 2007) to accommodate over-saturated flows and the Type-2 to manage multi-agent tasks (Balaji & Srinivasan, 2011). Taking the advantage of FL and reinforcement learning algorithms simultaneously, Fuzzy Q-learning is of interest to tune fuzzy inference parameters for each fuzzified state using Q-learning, related works can be found in (Moghaddam, Hosseini, & Safabakhsh, 2015) and (Moghaddam, Hosseini, & Safabakhsh, 2019).

**Genetic Algorithm (GA)** mimics the evolution process in biology, based on the philosophy of "Survival of the fittest" (Goldberg & Holland, 1988). This mechanism is transferred to TSC to search for optimal or near-optimal timing schedules, following the initialize--fitness computation--population evolution procedures (Foy and Benekohal 1992). Different variants of GAs find a broad range of applications in TSC, Memon & Bullen (1996) proposed a Generational Genetic Algorithm (GGA) to execute signal control in the LOCAL platform developed by the University of Pittsburg, and simulation indicates the optimization process is quicker and smoother than the Quasi-Newton model. GGA based control schemes can also be verified in (Park, 1998) and (Yun & Park, 2012). Aggregating GA and Q-learning is one trend in recent studies, GA plays an essential role in achieving optimal hyperparameters with quicker search speed, which enhances the real-time performance of traffic-response signal control. Coupled with Q-learning, a hierarchical-based multi-agent GA with a dynamic model is proven to reduce more delay than the classical GA (M. K. Tan, Chuo, Chin, Yeo, &

Teo, 2019). Readers can find a review on GA-based TSC algorithms in (Kesur, 2009).

**Reinforcement Learning (RL)**, one of the three basic machine learning paradigms, and its deep structured form, deep reinforcement learning (DRL), are the state-of-art AI scheme in TSC. RL studies optimal control algorithms where the interaction between agents and the environment can be modeled as a Markov Decision Process. In the context of TSC, the traffic system of a regional network is treated as the environment, each signal group at an intersection is abstracted to an agent, which executes action (a signal timing plan) $a_t$ at time step $t$, given the current state $s_t$, according to a policy $\pi$, then the environment gives feedback by sending a reward signal $r_t$ (see **Fig.5**). Agents use these experiences to polish their policies with the intention of maximizing the discounted cumulative returns. In settings of road networks, where each intersection cannot be considered as an isolated agent, multi-agent reinforcement learning schemes are necessary to model agents' cooperative and competitive behaviors. Readers can find more detailed reviews on theories and specific applications of RL in Chapter 2-4.

## 1.3 Thesis Outline and Contributions

**Chapter 2** introduces the basic knowledge of Markov Decision Process and the necessary formulation details of value-based deep reinforcement learning. **Chapter 3** models the coordination between network-level intersections as a Markov Game, extending the underlying mechanism to multi-agent reinforcement learning (MARL). With no need for global or regional centralized control, a family of fully decentralized MARL schemes is proposed in **Chapter 4**, together with specific definitions of the Markov objects used in the context of TSC. In **Chapter 5**, I illustrate the experiment configurations for a 4×4 grid network and discuss the results in detail. To close the gap between simulation and real-world control, in **Chapter 6**, I propose a policy-based RL scheme for future work.

The contribution of this thesis is three-fold,

1. Formulated the network-level traffic control problem as partially observable Markov Game.

2. Designed a family of fully decentralized deep reinforcement learning algorithms for multi-agent scenarios, which is shown to effectively reduce travel delay and dissipate waiting queues.

3. Proposed a conceptual policy-based RL scheme to close the gap between simulation and reality.

# 2 Deep Reinforcement Learning

Reinforcement learning (RL) can be used to solve optimal control problems where interactions between agents and the environment can be modeled as a Markov Decision Process (MDP). This chapter introduces the basic theories of RL and its deep structured form (with function approximators), which is called deep reinforcement learning (DRL).

## 2.1 Markov Decision Process

A Markov Decision Process (MDP) is a fundamental mathematical object that lies at the heart of most RL algorithms. The trial-and-error sequence learning process of RL can be modeled as an MDP, where the state transition and decision-making logic satisfy the Markov Property. According to the theory of Markov Chain, a temporal process can be simplified to a 2-tuple object $\mathcal{M} = \{\mathcal{S}, \mathcal{T}\}$, where

$\mathcal{S}$ is the state space, can be either continuous or discrete and states $s \in \mathcal{S}$

$\mathcal{T}$ is a transition operator defining the transition probability $p(s_{t+1} | s_t)$ of entering $s_{t+1}$ given the current state $s_t$. If not specified, $t$ refers to the control time step in this thesis.

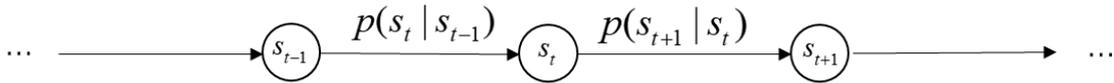

**Figure 2: Markov Chain**

In the control context, a Markov Chain is extended to a Markov Decision Process by introducing actions and rewards, therefore the 2-tuple object is enriched to $\mathcal{M} = \{\mathcal{S}, \mathcal{A}, \mathcal{R}, \mathcal{T}\}$, where

$\mathcal{A}$ is the action space, can be either continuous or discrete and actions $a \in \mathcal{A}$

$\mathcal{R} : \mathcal{S} \times \mathcal{A} \rightarrow \mathbb{R}$ is a reward function, which maps the $(s_t, a_t)$ vector to a real scalar $r_t$, causing instantaneous feedback if taking action $a_t$ under state $s_t$.

The Markov Property can be expressed as $p(s_{t+1} | s_t, a_t, r_t, s_{t-1}, a_{t-1}..., s_1, a_1, r_1) = p(s_{t+1} | s_t, a_t)$, the probability of moving from $s_t$ to $s_{t+1}$ just depends on the current state and action, regardless of which

trajectory it follows in the past.

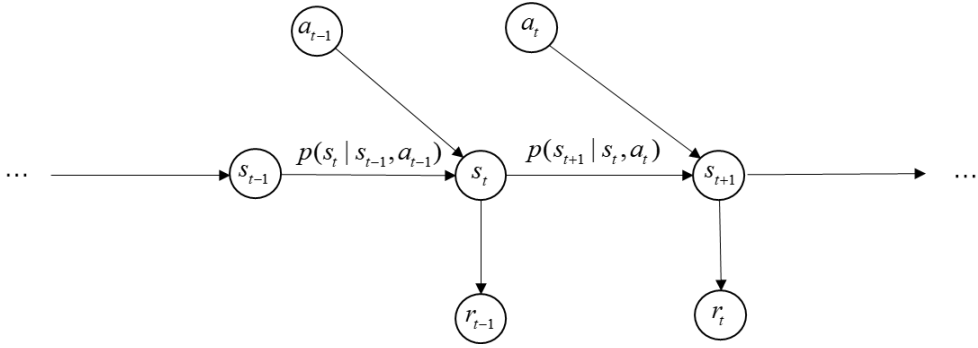

**Figure 3: Markov Decision Process**

However, in reality, an agent can hardly sense and understand states of the environment completely, in another word, the system is not fully observable. In that case, a Partially Observable Markov Decision Process (POMDP) is generated based entirely on observations, now the 4-tuple Markov object is further extended to $\mathcal{M} = \{\mathcal{S}, \mathcal{A}, \mathcal{O}, \mathcal{R}, \mathcal{E}, \mathcal{T}\}$, where

$\mathcal{O}$ is the observation space, can be either continuous or discrete and observations $o \in \mathcal{O}$

$\mathcal{E}$ is an omission probability distribution that $p(o_t | s_t) \sim \mathcal{E}$

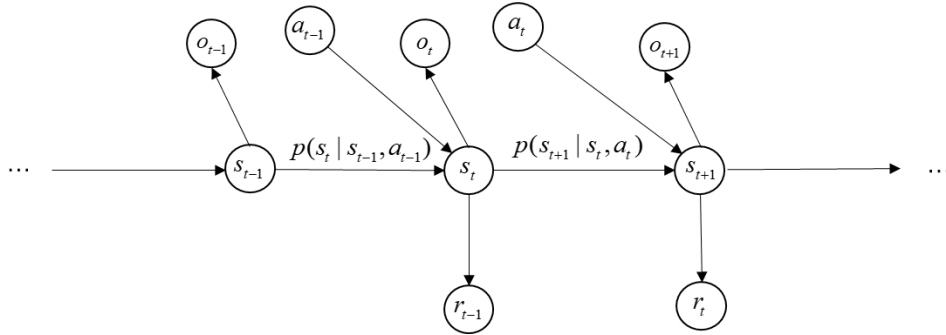

**Figure 4: Partially Observed Markov Decision Process**

Solving POMDPs optimally is computationally difficult when the system is beyond limited factors. One example of such concern arises when applying either proportional-integral-derivative(PID) or reinforcement learning algorithms to robotic control: in classical feedback control, torque, steering angles of arms and other measurable features can be extracted as control variables, while robots' sensors tend to omit features out of its vicinity due to limitations of sensing range and accuracy. Many efforts have been tried to better understand POMDPs, in this thesis, I use the decentralized method (Oliehoek, 2012) to tackle this problem.

## 2.2 Reinforcement Learning

Reinforcement learning is an area of machine learning that specifies in control issues, it differs from supervised learning and unsupervised learning, the two basic machine learning paradigms, in collecting samples while interacting with the environment. Rather than summarizing or giving prediction from labeled data, RL learns a reliable control strategy through a multistage decision-making process, which is often formulated as an MDP. This strategy, or more used as policy, maps states to actions, so as to maximize expected returns over time.

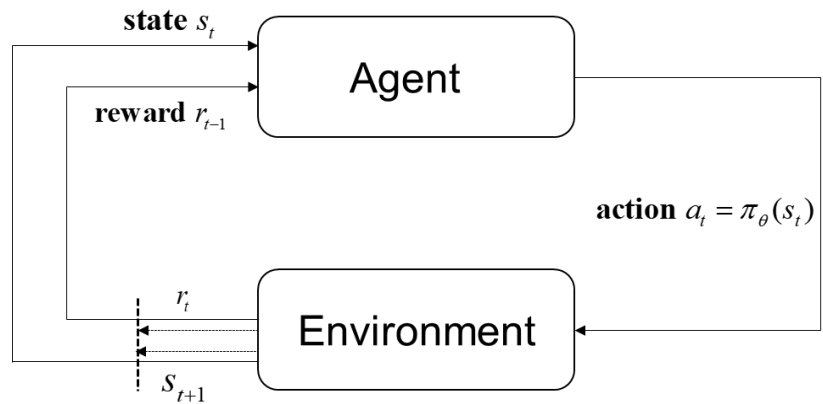

**Figure 5: Flow Chart of Reinforcement Learning**

**Fig. 5** illustrates how RL works: an agent perceives the current state of the environment $s_t$ at time step $t$, then executes an action $a_t$ under the instruction of a specific policy $\pi$ (parameterized as $\pi_\theta$ here). The loop continues as the environment receives the signal from the agent and gives a time-delay reward $r_t$ back. Simultaneously, the intrinsic state jumps from $s_t$ to $s_{t+1}$ with a probability of $p(s_{t+1} | s_t, a_t)$. In partially observable systems, agents interact with the world following POMDPs, $s_t$ hence degrades to $o_t$. We should note that the policy can be either good or bad here, depending on what criterion it takes. How to optimize a policy to get the best control performance under a given criterion is the core problem in RL.

The most used goal in RL is to maximize the expected cumulative reward $\bar{R}_\theta$ over finite time $T$ or an infinite time horizon, which is formulated as

$$\bar{R}_\theta = E_{\tau \sim p_\theta(\tau)}[\sum_t r(s_t, a_t)] \tag{2.1}$$

where $\tau$ is the state-action trajectory the agent follows, which can be expanded to $t = \{s_1, a_1, s_2, a_2, ..., s_t, a_t, ...\}$.

$p_q(t)$ is the joint probability distribution of $\tau$, short for

$$p_\theta(\tau) = p(s_1) \prod_{t=1}^{T} p_\theta(a_t | s_t) p(s_{t+1} | s_t, a_t) \quad \text{(finite time)} \tag{2.2}$$

$$p_\theta(\tau) = p(s_1) \prod_{t=1}^{\infty} p_\theta(a_t | s_t) p(s_{t+1} | s_t, a_t) \quad \text{(infinite time)} \tag{2.3}$$

where $p(s_1)$ is the probability of an initial state occurring. The cumulative reward is written as an expected value here for in model-free settings, the state transition probability $p(s_{t+1} | s_t, a_t)$ is not deterministic, it could change in different training episodes. $\bar{R}_q$ and $p_q(t)$ are associated with a parameter $q$, suggesting a corresponding policy $p_q$ is called. The policy can be either a direct mapping or a value function (e.g. Q function) plus a selection mechanism (e.g., $\varepsilon - \text{greedy}$). Now the objective is to find a vector $q^*$ so that

$$\theta^* = \arg\max_\theta E_{\tau \sim p_\theta(\tau)}[\sum_t r(s_t, a_t)] \quad \theta \in \Theta \tag{2.4}$$

Practically, there are two main philosophies in solving the policy optimization problem, namely value-based and policy-based methods. In the scope of value-based RL, a value function is utilized to evaluate how good it is to take a certain action under a certain state. Differently, policy-based RL algorithms optimize policies (usually with function approximation) directly, updating the parameters by gradient ascent. Next in this chapter, I will first introduce Q-learning, a state-of-art value-based RL method.

## 2.3 Q-Learning

### 2.3.1 Value Function

A value function is an estimation of expected cumulative reward from the current step $t$ to the end of episodes $T$, if the time horizon is infinite ($T = \infty$), a discounted factor is applied, where temporally closer rewards are given greater weights. When controlling traffic signals at an intersection, general traffic conditions can be extracted as states, suppose we are able to fully observe the traffic by taking a panoramic photo there, then with a little domain knowledge, one can expect better operation performance with low vehicle arriving rates than when the traffic demand is high, no matter what TSC strategy runs. Thus, the expected value function of the state $s_t$ is only related to the transition probability $p(s_{t+1} | s_t)$ and,

$$V(s_t) = \sum_{t'=t}^{T} E[r_{t'} | s = s_t] \tag{2.5}$$

The goal of RL is to find the optimal policy to guide which action the agent shall take given a certain state. Naturally, specifying the value of state-action pairs is more approachable. For a policy $p$, the expected state-action value, also known as Q value, is expressed as,

$$Q^\pi(s_t, a_t) = \sum_{t'=t}^{T} E_\pi[\mathcal{R}(s_{t'}, a_{t'}) | s_t, a_t] \tag{2.6}$$

Typically, short-term rewards are given preference, back to the TSC example, moving the traffic smoothly in the next few cycles is more rewarding since the traffic demand may change rapidly. Therefore, a discounted reward is designed by adding exponentially decreasing weights,

$$Q^\pi(s_t, a_t) = \sum_{t'=t}^{T} E_\pi[\gamma^{t'-t} \mathcal{R}(s_{t'}, a_{t'}) | s_t, a_t] \tag{2.7}$$

Where $\gamma \in (0,1]$ is a discount factor. One should note that $r(s_t, a_t)$ is an instant return given by the environment directly (see **Fig 5**), then the Q function is reorganized as,

$$Q^\pi(s_t, a_t) = r(s_t, a_t) + \sum_{t'=t+1}^{T} E_\pi[\gamma^{t'-t} \mathcal{R}(s_{t'}, a_{t'}) | s_t, a_t] \tag{2.8}$$

which can be written as an implicit recursive form,

$$Q^\pi(s_t, a_t) = r(s_t, a_t) + \gamma E_{s_{t+1} \sim p(s_{t+1}|s_t, a_t)}[V^\pi(s_{t+1})] \tag{2.9}$$

where $V^\pi(s_t) = \sum_{a_t \in \mathcal{A}} p_\pi(a_t | s_t) Q^\pi(s_t, a_t)$, is the expected return when starting at the state $s_t$ and following policy $\pi$ henceforth.

*Optimal Policy*: $p^*$ is the optimal policy, if $V^{\pi^*}(s) \geq V^\pi(s), \forall s \in \mathcal{S}$, for any other policy $\pi$, this optimal policy can be retrieved from a candidate policy $p_q$ by choosing actions greedily at each state:

$$\arg\max_{a \in \mathcal{A}} Q^{\pi_\theta}(s, a)$$

*Proof:*

$$\begin{aligned}
V^{\pi_\theta}(s) &= \sum_{a \in \mathcal{A}} p_\theta(a | s) Q^{\pi_\theta}(s, a) \leq \max_{a \in \mathcal{A}} Q^{\pi_\theta}(s, a) = Q^{\pi_\theta}(s, \pi^*(s)) \\
&= E_{\pi_\theta}[r_t + \gamma V^{\pi_\theta}(s_{t+1}) | s_t = s, a_t = \pi^*(s_t)] \\
&\leq E_{\pi_\theta}[r_t + \gamma Q^{\pi_\theta}(s_{t+1}, \pi^*(s_{t+1})) | s_t = s, a_t = \pi^*(s_t)] \\
&= E_{\pi_\theta}[r_t + \gamma r_{t+1} + \gamma^2 V^{\pi_\theta}(s_{t+2}) | s_t = s, a_t = \pi^*(s_t)] \\
&\leq E_{\pi_\theta}[r_t + \gamma r_{t+1} + \gamma^2 Q^{\pi_\theta}(s_{t+2}, \pi^*(s_{t+2})) | s_t = s, a_t = \pi^*(s_t)] \\
&= ... \leq V^{\pi^*}(s)
\end{aligned} \quad (2.10)$$

Where $p^*(s)$ is a deterministic action selection given state $s$, thus the optimal state-value function denoted $V^*(s)$ is defined as,

$$\begin{aligned}
V^*(s) &= \max_{a \in \mathcal{A}} E_{s_{t+1} \sim p(s_{t+1}|s_t, a_t)}[r_t + \gamma V^*(s_{t+1}) | s_t = s, a_t = a] \\
&= \max_{a \in \mathcal{A}} \sum_{s_{t+1}, r_t} p(s_{t+1}, r_t | s, a)[r_t + \gamma V^*(s_{t+1})]
\end{aligned} \quad (2.11)$$

The first and the second lines of equation (2.11) are two forms of the Bellman optimality equation for $V^*(s)$ (Bellman, 1957), accordingly, the Bellman optimality equation for $Q^*(s, a)$ is,

$$Q^*(s, a) = \sum_{s_{t+1}, r_t} p(s_{t+1}, r_t | s, a)[r_t + \gamma \max_{a_{t+1} \in \mathcal{A}} Q^*(s_{t+1}, a_{t+1})] \quad (2.12)$$

This equation states that the optimal value of taking *a* at *s* is the immediate reward plus the expected discounted optimal value attainable from the next state. If the environmental dynamics $(\mathcal{T}, \mathcal{R})$ are completely known, the Bellman optimality equation can be solved by Dynamic Programming or Asynchronous Dynamic Programming. While in most realistic systems, the state transition probability and how the reward function shapes are not explicit. To fill the gap, there are two main methodologies in understanding the inner mechanism of the system. One is model-based RL (Doya et al, 2002), where agents sample from interactions with the environment to infer $\mathcal{T}$ and $\mathcal{R}$, making it feasible to solve Bellman optimality equations with planning algorithms. The other one is model-free methods, which use an iterative approximation to estimate Q values from experience without a specific model. Model-free methods are widely used in RL research nowadays for in most cases, an agent won't know what is the next state and reward before it takes each action. Without a special explanation, the RL algorithms mentioned hereinafter are model-free.

The process of an RL algorithm is simplified to a 3-stage flow path, as shown in **Fig. 6**, the challenge is how to fit a model to estimate the value function or Q function so that it can exactly evaluate an action. Besides the known reward $r(s_t, a_t)$, the remainder of the right part of equation (2.9) is an expectation over all the random trajectories, it's technically hard to give a precise value with limited samples, variance is a critical issue. There are two main frameworks for learning Q values of state-action pairs, Monte-Carlo and Temporal Difference methods. They are designed to solve episodic ($T \neq \infty$) and non-episodic ($T = \infty$) MDPs correspondingly, combinations of these two methods also achieve impressive results (Peng & Williams, 1996).

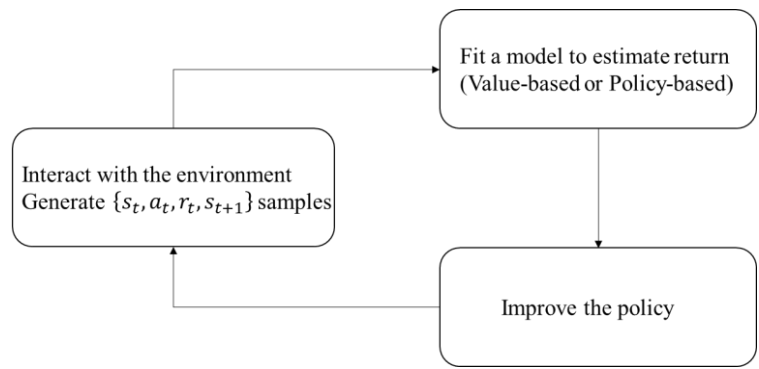

**Figure 6: The 3-stage Process of Reinforcement Learning**

### 2.3.2 Monte-Carlo Q Value Approximation

Monte-Carlo (MC) methods are ways of estimating value functions and discover optimal policies based on average sample returns, which require no prior knowledge. To ensure the availability of complete returns of a learning process, MC methods are constrained to episodic MDPs, where all episodes terminate to final states in finite time. A uniform policy is applied to all the time steps in an episode, only on the completion of each episode are value functions and thus policies changed.

We begin by estimating the state-action value function for a given policy $p$, not necessarily the optimal policy $p^*$. Each occurrence of a particular state-action pair $(s,a)$ is called a visit. There could be several visits to the same $(s,a)$ in an episode, and the first time it is covered is called the *first visit*. When sampling, the trajectory of state-action together with the instant reward $\{s_1,a_1,r_1,s_2,a_2,...,s_t,a_t,r_t,s_{t+1},...,s_T,a_T,r_T\}$ is stored in the memory. Right after the sampling process terminates at time $T$, this memory will be retrieved to calculate the long-term return $G_t(s_t,a_t)$ of taking $a_t$ at $s_t$, defined as,

$$G_t(s_t,a_t) = \sum_{t'=t}^{T} \gamma^{t'-t} \mathcal{R}(s_t,a_t) \qquad (2.13)$$

It is noteworthy that if the agent visits the same state $s$ at different times $t_1$ and $t_2$ ($t_1 \neq t_2$), and takes the same action $a$, then contradiction occurs whether we should calculate the cumulative discounted reward from $t_1$ or $t_2$. There are two MC algorithms to fix this multi-visit problem. The first-visit MC estimates $Q^p(s,a)$ as the average return following the first time in each episode that the state-action pair is visited. While the average-visit MC averages the $G(s,a)$ value of all the visits to $(s,a)$. For both two methods, as more time steps are experienced (the number of visits to each state-action pair approaches infinity), the average $G(s,a)$ converges to the true Q function of $(s,a)$ (R. S. Sutton & Barto, 2017a).

---

**Algorithm 1** *First-Visit* Monte-Carlo State-Action Value Estimation

**Input:** policy $\pi$, integer $num\_episodes$, integer $T(total\ time\ steps\ of\ an\ episode)$
**Output:** Value function $Q^\pi$
  initialize: Set $N(s,a) = 0$, $Returns(s,a) = 0$, $\forall s \in \mathcal{S}, a \in \mathcal{A}$
  **for** $i \leftarrow 1\ to\ num\_episodes$ **do**
    Generate a sequence $s_1, a_1, r_1, s_2, \dots\ s_t, a_t, r_t, s_{t+1} \dots\ s_T$
    **for** $t \leftarrow 1\ to\ T$ **do**
      **if** $(s_t, a_t)$ *is a first-visit* **then**
        $N(s_t, a_t) \leftarrow N(s_t, a_t) + 1$
        $Returns(s_t, a_t) \leftarrow Returns(s_t, a_t) + G_t(s_t, a_t)$
      **end if**
    **end for**
  **end for**
  $Q^\pi(s,a) \leftarrow Returns(s,a)/N(s,a)\ \forall s \in \mathcal{S}, a \in \mathcal{A}$
  Return $Q^\pi$

**Algorithm 2** *Every-Visit* Monte-Carlo State-Action Value Estimation
---
**Input:** policy $\pi$, integer $num\_episodes$, integer $T(total\ time\ steps\ of\ an\ episode)$
**Output:** Value function $Q^\pi$
  **initialize**: Set $N(s,a) = 0,\ Returns(s,a) = 0,\ \forall s \in \mathcal{S}, a \in \mathcal{A}$
  **for** $i \leftarrow 1\ to\ num\_episodes$ **do**
    Generate a sequence $s_1, a_1, r_1, s_2, \ldots s_t, a_t, r_t, s_{t+1} \ldots s_T$
    **for** $t \leftarrow 1\ to\ T$ **do**
      **if** $(s_t, a_t)$ is visisted **then**
        $N(s_t, a_t) \leftarrow N(s_t, a_t) + 1$
        $Returns(s_t, a_t) \leftarrow Returns(s_t, a_t) + G_t(s_t, a_t)$
      **end if**
    **end for**
  **end for**
  $Q^\pi(s,a) \leftarrow Returns(s,a)/N(s,a)\ \forall s \in \mathcal{S}, a \in \mathcal{A}$
Return $Q^\pi$
---

**Algorithm** 1 and 2 show the general procedures of Q function estimation for a policy, now we turn to use MC methods to approximate the state-action value function $Q^*$ of the optimal policy $p^*$. According to Generalized Policy Iteration (GPI), if policy evaluation (estimation of $Q^p$) and improvement ($\pi(s) = \arg\max_{a \in \mathcal{A}} Q^\pi(s,a)$) take place alternatively, an arbitrary policy $p_0$ is guaranteed to converge to $p^*$ (Sutton & Barto, 2017), where $E$ means "evaluation", $I$ means "improvement". The policy $p_k$ ($k = 1,2,3\ldots$) is a candidate for the optimal policy $p^*$, if $\pi_k$ selects action greedily: $\pi_k(s) = \arg\max_{a \in \mathcal{A}} Q^{\pi_{k-1}}(s,a)\ \forall s \in \mathcal{S}$, and $p_k$ is the optimal policy, if $\pi_k(s) = \pi^*(s)\ \forall s \in \mathcal{S}$.

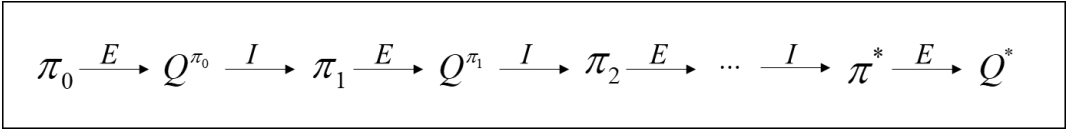

**Figure 7: Monte-Carlo Optimal Policy Iteration**

There is a critical issue in MC policy iteration that many state-action pairs may never be visited since candidate optimal policies always choose actions maximizing corresponding Q values. For those "ignored" actions, their Q values are hard to estimate thence won't be taken in iterations to go, even if in fact they are better. This *exploit-exploration* problem is essential in MC control, which requires every pair has a nonzero probability of being selected as the start. Different approaches have gained a good trade-off between exploitation and exploration (S. B. Thrun, 1992). In this thesis, $\varepsilon$ – greedy is the default technique to maintain exploration,

$$a = \begin{cases} \arg\max_{a \in \mathcal{A}} Q^{\pi_k}(s,a) & p = 1-\varepsilon \\ \text{random} & p = \varepsilon \end{cases} \quad (2.14)$$

The first-visit Monte-Carlo optimal control, with $\varepsilon$ – greedy is given below,

---

**Algorithm 3** *First-Visit* Monte-Carlo Optimal Control
---
**Input:** $\epsilon \in [0, 1)$, integer $num\_iteration$, integer $num\_episodes$, integer $T$
**Output:** Approximate optimal policy $\pi^*$
  initialize: Set $Q(s,a) = 0, \forall s \in \mathcal{S}, a \in \mathcal{A}$
  for $k \leftarrow 1$ to $num\_iteration$ do
    Set $N(s,a) = 0, Returns(s,a) = 0, \forall s \in \mathcal{S}, a \in \mathcal{A}$
    for $i \leftarrow 1$ to $num\_episodes$ do
      $\pi \leftarrow \epsilon - greedy(Q)$
      Generate a sequence $s_1, a_1, r_1, s_2, \ldots s_t, a_t, r_t, s_{t+1} \ldots s_T$
      for $t \leftarrow 1$ to $T$ do
        if $(s_t, a_t)$ is a *first visit* then
          $N(s_t, a_t) \leftarrow N(s_t, a_t) + 1$
          $Returns(s_t, a_t) \leftarrow Returns(s_t, a_t) + G_t(s_t, a_t)$
        end if
      end for
    end for
    $Q(s,a) \leftarrow Returns(s,a)/N(s,a) \forall s \in \mathcal{S}, a \in \mathcal{A}$
  end for
Return $\pi$

---

### 2.3.3 Temporal Difference Q-learning

MC control is actually not efficient in practice, because only when all the episodes in an iteration and all the time steps in an episode are finished can we estimate the Q function. In contrast, Temporal Difference (TD) methods need to only wait until the next time step, then update the Q function immediately and revise the policy right after that. The simplest TD method updates state-action values each time step,

$$Q^\pi(s_t, a_t) \leftarrow Q^\pi(s_t, a_t) + \alpha[r_t + \gamma V^\pi(s_{t+1}) - Q^\pi(s_t, a_t)] \quad (2.15)$$

Where $r_t + \gamma V^p(s_{t+1})$ is the expected value of $Q^p(s_t, a_t)$, also known as the *TD target*, the difference between the target value and the current estimation $r_t + \gamma V^p(s_{t+1}) - Q^p(s_t, a_t)$ is called the *TD error*, this TD *one-step* (TD(0)) algorithm uses the error term, multiplied by a constant discount $\alpha \in [0,1]$ to modify $Q^p(s, a | s = s_t, a = a_t)$. The state value function $V^p(s_{t+1})$ of $s_{t+1}$ is the expectation of $Q^p(s_{t+1}, a_{t+1})$, which can be approximated by samples from a deterministic policy.

The first principal TD control algorithm is SARSA (Andrew, 1998), named after the fact that it uses the quintuple of events, $\{s_t, a_t, r_t, s_{t+1}, a_{t+1}\}$, to approximate $Q^\pi(s_{t+1}, a_{t+1})$, therefore to update $Q^\pi(s_t, a_t)$,

$$Q^\pi(s_t, a_t) \leftarrow Q^\pi(s_t, a_t) + \alpha[r_t + \gamma Q^\pi(s_{t+1}, a_{t+1}) - Q^\pi(s_t, a_t)] \tag{2.16}$$

In SARSA optimal control, policy iterations are carried out every time step. SARSA is also called an *on-policy* method because it uses the same policy (not necessarily the greedy one) to take action at $t$, and predict $a_{t+1}$ at $t+1$, based on the observation of $s_{t+1}$.

---
**Algorithm 4** SARSA Optimal Control
---
**Input:** $\alpha \in [0, 1]$, $\epsilon \in [0, 1)$, integer $num\_episodes$
**Output:** Approximate optimal state-action value function $Q^*$
  initialize: Set $Q(s, a) = 0$, $\forall s \in \mathcal{S}, a \in \mathcal{A}$
  for $i \leftarrow 1$ to $num\_episodes$ do
    Observe the initial state $s_0$
    Take action $a_0$ according to $\epsilon - greedy(Q)$
      $t \leftarrow 0$
      while $s_t$ is not terminated do
        Take action $a_t$, and observe $r_t$ and $s_{t+1}$
        Choose action $a_{t+1}$, using $\epsilon - greedy(Q)$
        $Q(s_t, a_t) \leftarrow Q(s_t, a_t) + \alpha(r_t + \gamma Q(s_{t+1}, a_{t+1}) - Q(s_t, a_t))$
        $t \leftarrow t + 1$
      end while
  end for
Return $Q^*$
---

Another type of TD control algorithm is SARSA-max, as known as Q-learning (Watkins & Dayan, 1992). This breakthrough of off-policy learning simplifies the original SARSA and accelerates the converging speed. Instead of holding the assumption that the agent takes action following the current policy, Q-learning directly uses the maximum Q value of $s_{t+1}$ as the expected $Q(s_{t+1}, a_{t+1})$, that is we compulsively believe the agent will forward greedily (not $\varepsilon - greedy$).

---
**Algorithm 5** Q-learning
---
**Input:** $\alpha \in [0, 1]$, $\epsilon \in [0, 1)$, integer $num\_episodes$
**Output:** Approximate optimal state-action value function $Q^*$
  initialize: Set $Q(s, a) = 0$, $\forall s \in \mathcal{S}, a \in \mathcal{A}$
  for $i \leftarrow 1$ to $num\_episodes$ do
    Observe the initial state $s_0$
    $t \leftarrow 0$
    while $s_t$ is not terminated do
      Take action $a_t$ using $\epsilon - greedy(Q), observe$ $r_t$ and $s_{t+1}$
      $Q(s_t, a_t) \leftarrow Q(s_t, a_t) + \alpha(r_t + \gamma max_a Q(s_{t+1}, a) - Q(s_t, a_t))$
      $t \leftarrow t + 1$
    end while
  end for
Return $Q^*$
---

It's easy to prove that if SARSA involves a uniform maximization scheme, using the greedy policy rather than $\varepsilon-greedy$ to predict $a_{t+1}$, it will be identical with Q-learning. The convergence of Q-learning is guaranteed, a proof is available in (Melo, 2001). There are more extensions of SARSA and Q-learning, like Expected SARSA, Double Q-learning. For an overview of TD based reinforcement learning, I refer the readers to (Tesauro, 1992).

## 2.4 Value Function Approximation

Previous methodologies of Q-learning and other MC or TD algorithms are all developed based on a default assumption that the state space is discrete and not enormous. In this case, a computationally finite matrix ($|\mathcal{S}|\times|\mathcal{A}|$) is able to store Q values of each state-action pair, and exploits the greedy policies in a tabular manner, naturally, it's called Tabular Q-learning. However, this tabular method fails in many tasks where the state space is pretty large (e.g. absolute queue length of each lane as states in TSC of a big network) or not discrete such that the same state hardly appears twice (e.g. screenshots as states in Atari Game). Manually abstracting features to fit the table size also introduces extra issues of POMDP, like the problem of Hidden Markov Models (HMM), where the underlying environment is assumed to be Markovian, but the observations appearing to agents are not (Michael & Jordan, 1995).

To address the problem with memory needed for large discrete state spaces and value estimation accuracy for continuous state spaces, Q-value approximation methods are proposed. A good function estimator is required not only to exactly reflect the relative advantages of state-action pairs of current sampled tuples, but also to be able to generalize with states subject to the same distribution while never seen before. Much like a regression problem, function approximation is an instance of supervised learning, and thus extensive studies in the area of machine learning (ML), deep learning (DL) or other artificial intelligence (AI) branches have provided mature solutions for it. At this point, Q learning with function approximation is the coupling of conventional Q learning schemes (tabular value search) with state-action value function approximation, the latter one of which can be studied separately in this section.

Typically, an estimator of $Q^*(s,a)$ is denoted as $\hat{q}(s,a;\boldsymbol{\theta})$, where $\boldsymbol{\theta}$ is a weight vector or matrix parameterizing a specific function. The simplest and most intuitive form of this input-output mapping is linear regression,

$$\hat{q}(s, a_j, \boldsymbol{\theta}) = \boldsymbol{\theta}_j^T X(s) = \sum_{i=1}^{d} \theta_{ij} x_i(s) \qquad (2.17)$$

Where $a_j$ is the *j*-th ($j = 1, 2, ..., |\mathcal{A}|$) possible action (we just talk about the discrete action space here), $X(s)$ is called a *d*-dimensional *feature vector* that measures state *s*, with component $x_i : s \rightarrow \mathbb{R}$, these *basic functions* are designed with feature engineering from empirical studies or expert experience. Classical feature construction techniques are Polynomials (Kira & Rendell, 1992), Fourier Basis (Derrode & Ghorbel, 2001), Coarse Coding (R. R. S. Sutton, 1996), Tile Coding (Edgar An, Miller, & Parks, 1991), etc. Owing to good guarantees of convergence, the majority of works on RL with function approximation focused on linear approximators (Tsitsiklis & Van Roy, 1997), until the great success of using Deep Q Network to beat linear methods in playing Atari games (Mnih, Silver, & Riedmiller, 2013).

## 2.5 Deep Q-learning

Deep learning (DL) algorithms, which have outperformed many other numerical methods in pattern recognition, image classification, natural language processing, etc, are believed to be remarkably effective in approximating value functions in RL. The breakthrough of backpropagation (BP) techniques enables deep structured neural networks (DNNs) to process natural data in their raw form, without requiring manual feature engineering (Lecun, Bengio, & Hinton, 2015). With nonlinear connections (e.g. ReLU) between different layers (Dahl, Sainath, & Hinton, 2013), deep reinforcement learning (DRL) algorithms are making major advances in controlling complex systems, where the attempts of classical linear or nonlinear control have been thwarted. For a general review of DRL, I refer to (Arulkumaran et al., 2017) and (Li, 2018).

### 2.5.1 Deep Q Network

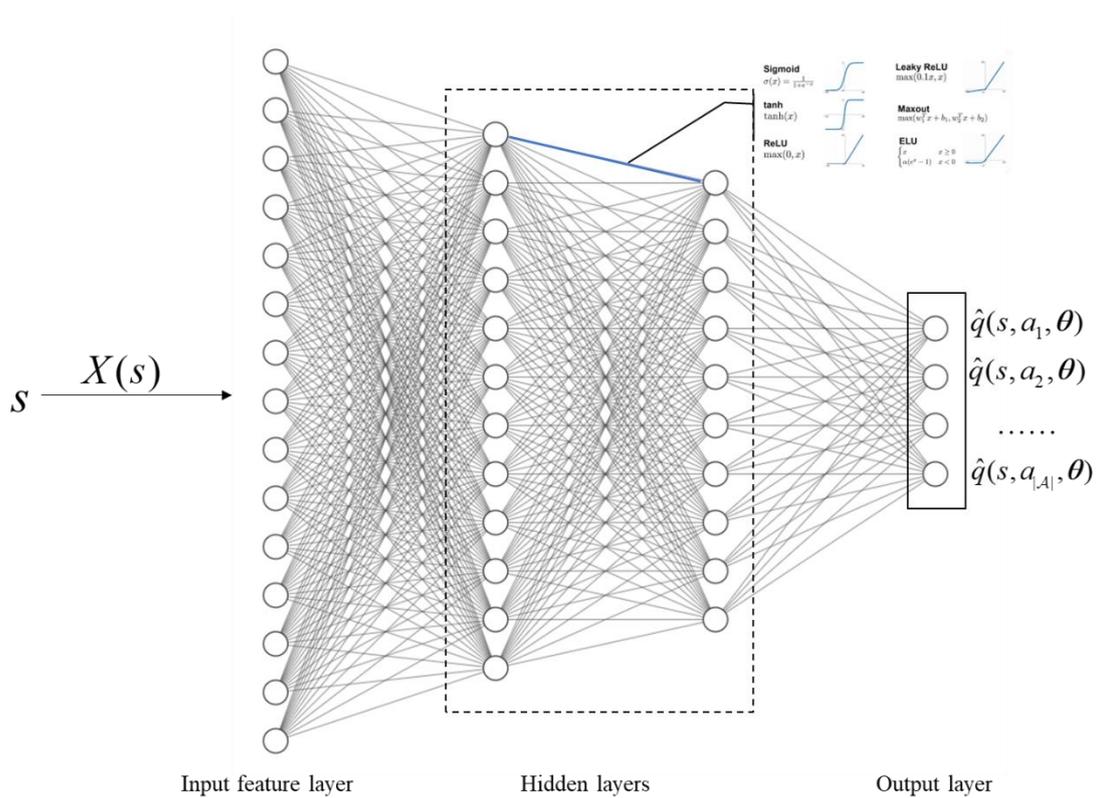

**Figure 8: Deep Q Network**

The priority of value-based RL algorithms is to estimate the value function, DNNs have been widely employed to tackle this task (Sallans & Hinton, 2004) (Maei et al., 2009), but not until the proposal of Neural Fitted Q-learning (NFQ) (Riedmiller, 2005) had researchers extended function approximation to Q-learning based control. Using stochastic gradient descent (SGD) (rather than the batch updates in NFQ) to update weights with experience replay, Deep Q Networks (DQN) (Mnih et al., 2013) outperforms all previous approaches on 6 of the 7 Atari 2600 games and surpassed a human expert on 3 of them. Mnih et al (2015) further combined the DQN algorithm with a convolutional neural network (CNN) (Krizhevsky, Sutskever, & Hinton, 2012). Rather than connecting all the layers fully in the original DQN, convolutional layers and pooling mechanisms (Nagi et al., 2011) in the advanced version of DQN enabled agents to process raw images without omitting information, and achieved human-level performance in playing 49 Atari games. Considering well-recognized advantages in nonlinear control, especially where the action space is discrete and not too large, I take DQN as the default RL algorithm in our TSC task, where the states are considerably complex but actions are simple as alternating red and green lights.

The optimal state-action value function estimator $\hat{q}(s,a;\boldsymbol{\theta})$ is represented as a multi-layer artificial neural network (see **Fig. 8**). Before fed into fully connected layers, features of inner states can be extracted by expert experience or CNNs (if the sensing technology is computer vision), this part is denoted as $X(s)$, for convenience, I still write it as $s$ in the remaining parts. To maintain the nonlinearity of the feedforward propagation and make derivative operations on weights feasible, a simple nonlinear function is set to connect different neurons. In this article, Rectified Linear Units (ReLUs) with a function $f(x) = \max(0, x)$ are implemented. The output layer of DQN is a real-value vector indicating the estimated optimal Q values of each $(s, a_i)\ i = 1, 2, ..., |\mathcal{A}|$.

To update network weights with the error backpropagation method (Hagan & Menhaj, 1994), an expected mean square error (MSE) loss function is defined,

$$L_k(\boldsymbol{\theta}) = E_{\tau \sim p_{\boldsymbol{\theta}}(\tau)}[(y_k - \hat{q}(s,a;\boldsymbol{\theta}))^2] \tag{2.18}$$

Where $y_k = E_{s' \sim p(s'|s,a)}[r(s,a) + \gamma \max_{a'} \hat{q}(s',a';\boldsymbol{\theta}) | s, a]$ is the target of $Q^*(s,a)$ at iteration $k$, the derivative with respect to $\boldsymbol{\theta}$ is written as the expected gradient form too,

$$\nabla_{\boldsymbol{\theta}} L_k(\boldsymbol{\theta}) = E_{\tau \sim p_{\boldsymbol{\theta}}(\tau), s' \sim p(s'|s,a)}[(r(s,a) + \gamma \max_{a'} \hat{q}(s',a';\boldsymbol{\theta}) - \hat{q}(s,a;\boldsymbol{\theta}))\nabla_{\boldsymbol{\theta}} \hat{q}(s,a;\boldsymbol{\theta})] \tag{2.19}$$

$\boldsymbol{\theta}_{k+1}$ is updated by $\boldsymbol{\theta}_{k+1} = \boldsymbol{\theta}_k - \eta \nabla_{\boldsymbol{\theta}} L_k(\boldsymbol{\theta})$, $\eta \in (0,1)$ denotes the learning rate. In practice, this expected value is technically hard to calculate since we are not able to collect all the possible $(s',a'|s,a,r)$ pairs given the state transformation probability is unknown. Similar to other supervised learning algorithms, instead of using the full training data, every time we randomly select a batch to derive the gradients,

$$\nabla_{\boldsymbol{\theta}} L_k(\boldsymbol{\theta}) = [r + \gamma \max_{a'} \hat{q}(s',a';\boldsymbol{\theta}) - \hat{q}(s,a;\boldsymbol{\theta})]\nabla_{\boldsymbol{\theta}} \hat{q}(s,a;\boldsymbol{\theta}) \tag{2.20}$$

It is noteworthy that different from supervised learning cases, the target value of DQN is involved with $\theta$, which will cause oscillations through the iterative learning process. Mnih et al. (2015) gave a *fixed target* solution to this problem: the parameters $\theta^-$ of the target network $\hat{q}(s',a';\theta^-)$ are only updated with the DQN parameters $\theta$ every *C* steps, and kept fixed during that period,

$$\nabla_{\theta} L_k(\theta) = [r + \gamma \max_{a'} \hat{q}(s',a';\theta^-) - \hat{q}(s,a;\theta)] \nabla_{\theta} \hat{q}(s,a;\theta)$$
$$\text{Every } C \text{ steps, } \theta^- \leftarrow \theta$$
(2.21)

Another breakthrough made by (Mnih et al., 2013) is the *Experience Replay* approach, an offline Q-learning scheme that improves data efficiency greatly over standard online Q-learning. At each time step *t*, the agent takes an action using $\varepsilon - \text{greedy}$, then a 4-tuple "experience" $e_t = \{s_t, a_t, r_t, s_{t+1}\}$ is stored in a memory $\mathcal{D}$ with size *N*, right after that, several samples $\{e_1^{(t+1)}, e_2^{(t+1)}, ..., e_b^{(t+1)}\}$ (*b* is the size of a minibatch) are pulled randomly from the memory to update $\theta$. Besides potentially using an experience more than once, experience replay has another advantage in decoupling correlations between consecutive samples. With the TSC task in mind, located at an intersection approached by an arterial road with high demand and a branch road with low demand, the undertrained signal tends to keep giving green to movements in the main road, because it receives continuous positive feedback and thus remains at a poor local maximum. As a result, vehicles heading other directions get stuck there. With more upstream flow arriving, the queues at side streets end up blocking neighboring intersections, causing the algorithm to diverge catastrophically. A typical DQN algorithm with fixed target network and experience replay is described as below,

**Algorithm 6** Deep Q-learning
---
**Input:** $\eta \in (0,1)$, $\epsilon \in [0,1)$, integer $num\_episodes$, integer $C$, memory size $N$, batch size $b$
**Output:** Optimal state-action value function $\hat{q}(s,a;\theta)$
    **initialize**: Set $\hat{q}(s,a;\theta)$ with random weights $\theta_0$
    **for** $i \leftarrow 1$ $to$ $num\_episodes$ **do**
        Observe the initial state $s_0$
        $t \leftarrow 0$
        **while** $s_t$ is not terminated **do**
            Take action $a_t$ using $\epsilon - greedy(\hat{q}(s_t,a;\theta))$, observe $r_t$ and $s_{t+1}$
            Store transition $e_t = \{s_t, a_t, r_t, s_{t+1}\}$ in $\mathcal{D}$
            Sample random minibatch $\{e_1^{(t+1)}, e_2^{(t+1)}, ..., e_b^{(t+1)}\}$ from $\mathcal{D}$
            **if** $t \% C == 0$ **then**
                $\theta^- \leftarrow \theta$
            **end if**
            **for** $j \leftarrow 1$ $to$ $b$ **do**
                $\{s,a,r,s'\} \leftarrow e_j^{(t+1)}$
                $d\theta \leftarrow d\theta - [r + \gamma max_{a'}\hat{q}(s',a';\theta^-) - \hat{q}(s,a;\theta)]\nabla_\theta \hat{q}(s,a;\theta)$
            **end for**
            $\theta \leftarrow \theta - \eta d\theta$
            $d\theta \leftarrow 0$
            $t \leftarrow t + 1$
        **end while**
    **end for**
    Return $\theta$

### 2.5.2 Towards Convergence and Stability

Deep Q Learning (DQL) has been achieving impressive success in high challenging tasks, e.g., robotic control (Gu et al, 2016), natural language processing (Sharma & Kaushik, 2017), finance (Deng et al., 2017), healthcare (Ling et al., 2017) and etc. While there still exists a gap between DQL theories and real-world operations, as function approximators tend to cause non-convergence and instability (Baird, 1995). Early-stage researches mainly focus on improving practical performance by applying modifications to neural network structures or Q learning procedures. Kick started by the original DQN algorithm, Prioritized Experience Replay (Schaul et al., 2015) adds replay weights to the uniform selection mechanism, replaying more often transitions with higher TD errors. As a deep version of Double Q-learning (Van Hasselt, 2010), Double DQN (DDQN; Van Hasselt, Guez, and Silver 2016) addresses the overestimation of Q values by decoupling selection and evaluation of optimal actions. The dueling network architecture (Wang et al., 2015) trains two separate estimators simultaneously, one for the state value function and one for the state-dependent action advantage function, which overperforms DDQN in playing Atari games. The variants of asynchronous methods for deep reinforcement learning are proposed in (Mnih et al., 2016), by running parallel interactions on a single machine with a standard multi-core CPU, the best performing algorithm, asynchronous advantage actor-critic (A3C) surpasses most state-of-art DQN algorithms in both playing Atari games and continuous motor control, with just half of the training time that replay-memory-required methods need. Distributional Q-learning (Bellemare, Dabney, & Munos, 2017) learns the distribution of estimated returns rather than the mean value previous algorithms suggest, this risk-aware tool gives us a sense of the instability in the Bellman optimality operator and the state aliasing problem even in a deterministic environment. Different from conventional $\varepsilon-$greedy methods, Noisy Network (Fortunato et al., 2017) is a novel action exploration mechanism, where parametric noise is added to neural networks' weights, thus to introduce stochasticity to agents' policies.

The DeepMind team combined all these improvements to an integration called Rainbow (Hessel et al., 2018), experiments show that different collocations achieve various performance, Distributional DQN and Dueling DDQN are the two most efficient ones, while A3C gains the least improvement compared with the raw DQN. Next in this section, Prioritized Experience Replay, Double DQN and the Multi-step Bootstrapping (R. S. Sutton & Barto, 2017) are introduced, as these techniques will be used as an augmentation to the original DQN in this thesis. For a theoretical understanding of DQN and statistical error analysis on convergence, I refer readers to (Yang, Xie, & Wang, 2019).

**Prioritized Experience Replay**  Experience replay (Lin, 1992) (Mnih et al., 2013) (Mnih et al., 2015) is the key technique to achieve stability in DQN, as it decouples the correlation between consecutive data by storing MDP trajectories to a memory and sampling a minibatch of them randomly during each update step. In addition, an experience will theoretically be picked up more than once, which definitely increases the data usage efficiency of DQN. Rather than iterate Q functions in a temporally continuous manner, experience replay allows the approximator to learn from a mixture of new and old data, this diversity contributes to the generalization ability for independent and identically distributed states.

Behaving in a fair way, experience replay selects experiences uniformly from the memory, that is common experiences that appear more often have greater possibilities to be chosen, while rare but potentially-with-high-information ones are therefore neglected. This "Matthew Effect" makes the original experience replay mechanism less efficient when the replay memory size is large.

A solution to the above problems is Prioritized Experience Replay (Schaul et al., 2015), it borrows the ideas from Prioritized sweeping (Moore & Atkeson, 1993), giving priority (greater probability to be selected) to experiences with higher TD error. This intuitive method assumes the Q network does not make good value approximation on these experiences as the TD errors haven't been eliminated to a satisfactory level, which calls for more frequent learning. Greedy TD-error prioritization replays the transition with the largest absolute TD error right after the most recent experience since the later one does not have a known TD error before training. However, this greedy method tends to just focus on a small subset of the experience memory, as TD errors go slowly when a nonlinear function approximator is used, as a result, information of samples with low TD error are therefore omitted. In addition, greedy prioritization is sensitive to outliers, noise spikes will dominate the replay process, feeding back wrong information again and again.

Similar to the Boltzmann machine (Crawford et al., 2018) used for stochastic action selection, a stochastic prioritization sampling mechanism is introduced (Schaul et al., 2015),

$$P(i) = \frac{p_i^\alpha}{\sum_k p_k^\alpha} \tag{2.22}$$

$P(i)$ is the sampling probability of experience $i$, where $p_i$ is the corresponding priority. The exponent $\alpha \geq 0$ measures the extent of greediness, $a = 0$ representing the uniform case. $p_i$ is calculated following either a proportional ($p_i = |d_i| + e$) or rank-based ($p_i = 1/\text{rank}(i)$) method, where $|\delta_i|$ is the TD error of experience $i$, $\varepsilon$ is a small, positive number preventing zero probability, $\text{rank}(i)$ is the rank of experience $i$ when the replay memory is sorted according to $|d_i|$. In algorithms hereinafter, the proportional prioritization is used. Besides, I refer readers to (Bruin et al., 2015) for replay memory structure design, and (Liu & Zou, 2019) for determining proper buffer sizes of replay memories.

**Double DQN** Estimating Q function with a maximization step, DQN tends to overestimate state-action values due to undertrained function approximation (S. Thrun & Schwartz, 1993) and environment noise (Van Hasselt, 2010). Van Hasselt et al. (2016) found, overoptimistic Q estimation, thus resulted poor polices, still exists even in deterministic settings, where harmful noise rare happens. Using the same network to select the best action and to evaluate the corresponding state-action pair is believed to be the main cause, the upper bound of overestimation is given in (S. Thrun & Schwartz, 1993). The fixed TD target $y_t = r_t + \gamma \max_a \hat{q}(s_{t+1}, a; \boldsymbol{\theta}^-)$ can be untangled to two procedures, i.e., selection and evaluation,

$$\hat{a}_{t+1} = \arg\max_a \hat{q}(s_{t+1}, a; \boldsymbol{\theta}^-) \tag{2.23}$$

$$y_t = r_t + \gamma \hat{q}(s_{t+1}, \hat{a}_{t+1}; \boldsymbol{\theta}^-) \tag{2.24}$$

Double DQN decouples them by using the local network to select the best action and applying the target network to evaluate its Q value, setting the TD target as,

$$y_t = r_t + \gamma \hat{q}(s_{t+1}, \arg\max_a \hat{q}(s_{t+1}, a; \boldsymbol{\theta}); \boldsymbol{\theta}^-) \tag{2.25}$$

In practice, $\boldsymbol{\theta}$ is updated every step, whereas $\boldsymbol{\theta}^-$ a fixed several-step interval.

**Multi-step Bootstrapping** Multi-step bootstrapping is more like a compromise lying between Monte-Carlo and one-step TD learning, which performs updates based on an intermediate number of experiences rather than just an instant one or all of them until termination. This n-step TD learning method is designed to eliminate the high variance of rewards caused by MC control, and the instability problem with one-step TD control. Parameters of the local network are updated every n step, during which the agent interacts with the environment using the same policy. An advanced version of Multi-step Double DQN with prioritized experience replay is shown in **algorithm 7**.

**Algorithm 7** n-step Double Deep Q-learning with Prioritized Experience Replay
---
**Input:** $\eta \in (0,1)$, $\epsilon \in [0,1)$, integer $num\_episodes$, integer $C$, TD step $n$, memory size $N$, batch size $b$
**Output:** Optimal state-action value function $\hat{q}(s,a;\theta)$
    **initialize**: Set $\hat{q}(s,a;\theta)$ with random weights $\theta_0$
    **for** $i \leftarrow 1$ *to num_episodes* **do**
        Observe the initial state $s_0$
        $t \leftarrow 0$
        **while** $s_t$ is not terminated **do**
            Take action $a_t, a_{t+1}, ... a_{t+n-1}$ using $\epsilon - greedy(\hat{q}(s,a;\theta))$, s=$s_t, s_{t+1}, ... s_{t+n-1}$
            Store transition $e_t, e_{t+1}, .., e_{t+n-1}$ in $\mathcal{D}$
            Sample random minibatch $\{e_1^{(t+n)}, e_2^{(t+n)}, ..., e_b^{(t+n)}\}$ from $\mathcal{D}$
            **if** t % C == 0 **then**
                $\theta^- \leftarrow \theta$
            **end if**
            **for** $j \leftarrow 1$ *to b* **do**
                $\{s,a,r,s'\} \leftarrow e_j^{(t+n)}$
                $d\theta \leftarrow d\theta - [r + \gamma max_{a'} \hat{q}(s',a';\theta^-) - \hat{q}(s,a;\theta)] \nabla_\theta \hat{q}(s,a;\theta)$
            **end for**
            $\theta \leftarrow \theta - \eta d\theta$
            $d\theta \leftarrow 0$
            $t \leftarrow t + n$
        **end while**
    **end for**
    Return $\theta$

# 3 Multi-agent Reinforcement Learning

Modeling traffic signal control as MDPs (Yu & Stubberud, 1997), different frameworks of Reinforcement Learning have been widely developed to achieve state-of-art performance in real-time manners. To the best knowledge of the author, the earliest research can be traced back to (Thorpe, 1997), where 4x4 identical intersections share common state-action values across the network while training their own sets of eligibility traces with tabular SARSA individually. Wiering (2000) applied a model-based RL scheme to infer state transition distributions, thus to accelerate the training process.

On the watershed when the DQN (Mnih et al., 2013) algorithm changed the game, El-Tantawy, Abdulhai, & Abdelgawad (2014) summarized pre-DQN RL methods in TSC, those algorithms mainly rely on discrete human-crafted state division and tabular state-action value search, which tends to omit important information and are hardly scalable to large networks. Research on advanced TSC catches the pace of DRL right after the proposal of variants of deep value-based or policy-based algorithms, most of which focus on improving the control performance of a single intersection. Gao et al. (2017) divide entrance lanes of the intersection to discrete cells, then integrates them into matrices of vehicle positions and velocities. Entries of the position matrix are filled in with 0/1 to indicate absence/ presence of vehicles on that road segment and the elements of the velocity matrix are normalized by the speed limit. These two matrices together with a current-action-flag vector are fed into a 6-layer DNN with a convolutional layer. Using an experience replay mechanism, simulation results in SUMO (Krajzewicz et al. 2012, will be introduced later) show this work reduces vehicle delay by 47% compared with the Longest Queue First algorithm. Liang et al. (2019) incorporates the original DQN with fashionable optimization techniques like the Dueling Network, Double Q-learning, and the Prioritized Experience Replay, this integration surpasses other comparison peers with just one or two of these optimization techniques. The first policy-based RL algorithm in the context of TSC was proposed by (Mousavi, Schukat, & Howley, 2017). Learning directly from snapshots of the intersection in SUMO, this policy-based method is reported to gain comparable performance with value-based tactics.

Wiering (2000) pointed out the importance of considering the cooperation mechanism between networked traffic nodes, rather than treating each signal as an isolated agent, which is believed to put the entire system under a sub-optimal situation. In the setting of multi-agent systems (MASs), reinforcement learning is therefore enriched to multi-agent reinforcement learning (MARL). Besides the tyranny of dimension already existing in single-agent RL problems, several new challenges arise for MARL. Foremost of these is the exponential growth of state-action space as the number of agents increases, this explosion of dimension unquestionably burdens computational complexity heavily. Secondly, specifying reward functions for agents is difficult, since their returns are highly correlated, each agent is affected by the others more or less. In the case of sparse reward signals, inverse reinforcement learning (IRL) is the mainstream algorithm to infer reward functions by learning from demonstrations (Natarajan et al., 2010). Thirdly, nonstationarity is unavoidable in MARL, much like the moving target problem in single-agent RL, the learned optimal policies for individual agents may possibly collapse, because, in the setting of decentralized partially observed Markov decision process (Dec-POMDP), distributions of agents' observations change over time as long as their local policies converge asynchronously.

This chapter is intended to introduce the concept of Markov Games for modeling multi-agent systems, and to give a brief formulation of Nash Policies which are targeted as common goals in MARL. Considering the diversity and complexity of agents' interactions, there are no unified theories that apply to general coordination scenarios, such models are usually developed in an ad hoc manner. Thus, I only review two basic communication mechanisms, i.e., sharing of network parameters and experience replay memories.

MARL is promising to fill the gap between RL research and real-world applications, where no one is isolated, theoretically. I refer readers to (Buşoniu, Babuška, & De Schutter, 2008) for a comprehensive overview of MARL, and to (J. N. Foerster, 2018) for deep learning based MARLs.

**3.1 Markov Game**

Markov game (MG), also known as Stochastic Game, is the extension of MDP in multi-agent environments, with agents interacting with each other to tackle cooperative or competitive tasks (Littman, 1994). In its general form, the ingredients of an MG is defined as a tuple $\left(\mathcal{S}, \{\mathcal{A}^i\}_{i \in N}, \{\mathcal{R}^i\}_{i \in N}, \mathcal{T}\right)$, where $N$ is the number of agents, $\mathcal{S}$ is the state space (measurable) of the entire environment. For agent $i$, $\mathcal{A}^i$ is the action space, $\mathcal{R}^i : \mathcal{S} \times \mathcal{A} \times \mathcal{S} \to \mathbb{R}$ is the local reward function. $\mathcal{T} : \mathcal{S} \times \mathcal{A} \times \mathcal{S} \to [0,1]$ is the global state transition probability distribution. One should notice here that $\mathcal{R}^i$ and $\mathcal{T}$ are dependent on the joint action space $\mathcal{A} = \mathcal{A}^1 \times \mathcal{A}^2 \times \cdots \mathcal{A}^N$ (I denote joint quantities of agents in bold), meaning the state transition is determined by conjunct actions of agents, and one's action influences the others' rewards directly or indirectly. Similarly, I define the local policy of agent $i$ as $\pi^{(i)} : \mathcal{S} \times \mathcal{A}^i \to [0,1]$, which maps a global state $s \in \mathcal{S}$ to a local action $a^i \in \mathcal{A}^i$ with a probability $\pi^{(i)}(a^i \mid s) \in [0,1]$.

To evaluate how well an agent responds to the dynamics of the environment, a distributed state-action value $Q^{p_i}(s, \boldsymbol{a})$ is designed to measure the cumulative discounted return of agent $i$, w.r.t $p^{(i)}$. For convenience, $Q^{p_i}$ is replaced by $Q^{(i)}$ hereinafter. $\{\mathcal{R}^i\}_{i \in N}$ is shaped by the synergism between agents. In a fully cooperative setting, all the reward functions are identical that $\mathcal{R}^1 = \mathcal{R}^2 = \cdots = \mathcal{R}^N$. The opposite is the fully competitive scenario, also called zero-sum MG, as $\mathcal{R}^1 + \mathcal{R}^2 + \cdots + \mathcal{R}^N = 0$, and literature on this field typically focuses on two-player cases (Vamvoudakis & Lewis, 2012).

Intermediate circumstances lying between the fully cooperative and competitive ones are called general-sum (or mixed) games (Lowe et al., 2017).

Markov Game involves both MDP and game theories, thus research progresses in these two lines together contribute to the development of MG. Readers can find related works in (Breen, 2017) and (Shoham & Leyton-Brown, 2008), I also refer to (Solan, 2012) for a detailed review on MG.

**3.2 Nash Policies**

Centralized MARL is exposed to many known problems, i.e., scalability and computational complexity, and in most cases can degrade to single-agent RL. For these reasons, hereinafter we refer MARL as decentralized MARL (Dec-MARL), where no central authority is responsible for allocating resources, and each agent has its own local policy and state-action functions. Intuitively, we borrow the idea of Q function from single-agent RL, which is defined as a recursive form of expected discounted return,

$$Q^{(i)}(s,\boldsymbol{a}) = \mathcal{R}^i(s,\boldsymbol{a}) + \gamma \sum_{s' \in S} \mathcal{T}(s,\boldsymbol{a},s') \cdot \sum_{a^{(1)'} \in \mathcal{A}^1, \ldots, a^{(N)'} \in \mathcal{A}^N} \prod_{i=1}^{N} \pi^{(i)}(a^{(i)'} \mid s') Q^{(i)}(s',\boldsymbol{a}') \qquad (3.1)$$

Where $\boldsymbol{a} = a^1 \times a^2 \times \cdots \times a^N$ is the joint action of agents. Each participant has its own Q function, though the corresponding reward functions are dependent on joint actions and the global state transition probability. Typically, it is unclear how to perfectly measure the performance of a MAS (Shoham, Powers, & Grenager, 2007). The boundary between cooperation and competition is sometimes ambiguous that even cooperative agents may have to be 'selfish' when their immediate returns are conflicting. An important and well-recognized criterion in MARL research is to reach *Nash Equilibria* (Ehtamo, 1997). Suppose all the policies except for $\pi^{(i)}$ are fixed, the best-response Q function $Q^*_{\Delta i}(s,a)$ (Littman, 2001) for agent *i* can be obtained by,

$$Q^*_{\Delta i}(s,\boldsymbol{a}) = \mathcal{R}^i(s,\boldsymbol{a}) + \gamma \sum_{s' \in S} \mathcal{T}(s,\boldsymbol{a},s') \cdot \max_{a^{(i)'} \in \mathcal{A}^i} \sum_{a^{(1)'},\ldots,a^{(i-1)'}} \sum_{a^{(i+1)'},\ldots,a^{(N)'}}$$
$$\times \pi^{(1)}(a^{(1)'} \mid s') \cdots \pi^{(i-1)}(a^{(i-1)'} \mid s') \cdot \pi^{(i+1)} \qquad (3.2)$$
$$\vee (a^{(i+1)'} \mid s') \cdots \pi^{(N)}(a^{(N)'} \mid s') Q^*_{\Delta i}(s',\boldsymbol{a}')$$

The idea here is agent *i* has an optimal policy $\pi^{(i)*}$ that achieves the best-response Q value in state s, when policies of the other agents are known,

$$E\{Q^{(i)} \mid \pi^{(1)},\ldots,\pi^{(i)},\ldots,\pi^{(N)}; s\} \leq E\{Q^{(i)} \mid \pi^{(1)},\ldots,\pi^{(i)*},\ldots,\pi^{(N)}; s\} \quad \forall \pi^{(i)} \qquad (3.3)$$

To make a more general case, if all the agents can take the best responses to the others, then a *Nash Equilibrium* is achieved, $[\pi^{(1)*}, \pi^{(2)*},\ldots,\pi^{(N)*}]$ are called *Nash policies*. Under the *Nash Equilibrium*, no agent can benefit more as long as the others all take *Nash Policies*. Converging to a *Nash Equilibrium* is the common goal in MARL, Filar & Vrieze (1996) proved that there at least exists a *Nash Equilibrium* for MGs in stationary policies but this conclusion is not guaranteed if agents take stochastic policies (i.e., the 'Rock, Paper, Scissor' game).

### 3.3 Communication for Cooperation

First of all, we define cooperative agents as those who are not able to get better equilibrium returns unless they use coordination mechanisms. In this part, I introduce two inter-agent information sharing philosophies and the communication effectuation supporting them.

**Parameter Sharing**    In deterministic policy settings, a critic exists to learn from experiences and to help actors pick the best action, which can be a state-action value table, a linear function approximator or a parameterized neural network. Sharing the parameters of such actors between agents is proven effective in improving the performance of deep distributed recurrent Q-networks (DDRQN) agents (J. N. Foerster et al., 2016).

Typically, parameter sharing is beneficial where cooperative agents shape and behave approximate homogenously. Take deep-structured MARL for example, agents sharing parameters usually have identical neural network structures (not necessarily the weights), at least some layers of them are the same, e.g., they all take 960 * 960 images as input or use the same pooling layer to extract features. Parameter sharing approaches can be the full copy between agents' critics or just a fixed number of layers.

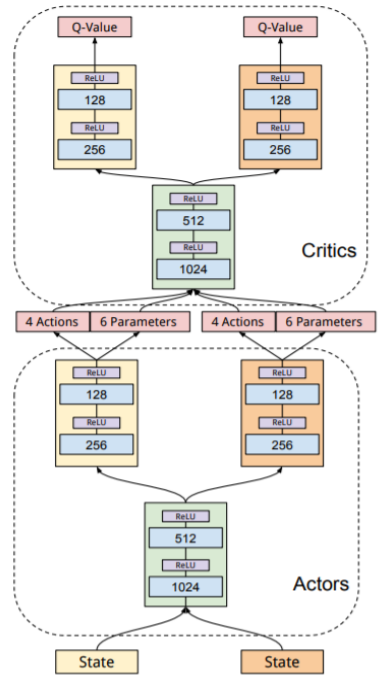

**Figure 9: Partially Parameter Sharing Architecture** (Hausknecht, 2016)

**Fig. 9** illustrates the partial parameter sharing approach proposed in (Hausknecht, 2016), where both two layers of actors and critics of two agents are shared. In his doctoral dissertation, Hausknecht also suggests skills that the parameters of the lower layers of networks are of more importance for sharing, since they are responsible for basic state processing and serve as the foundation for upper decisions. Sharing the lowest layers, agents still maintain the specification of higher layers, which allow for developing unique policies.

**Memory Sharing**     Besides sharing weights of network parameters, approximate homogenous agents have another choice to speed up the convergence, which is to share their SARSA trajectories to a common replay memory. This memory serves as a shared pool where agents can sample mixed experiences from either themselves or teammates. Much like the prioritized experience replay, this coordination technique adds extra diversity to the memory, making models more robust to generalize. However, how to learn from others' experiences is an open question in MARL, since the local environment and observation distributions of agents may differ.

# 4 S2R2L: A Decentralized Multi-agent Reinforcement Learning Strategy for Network Level Traffic Signal Control

To overcome the inherent weakness of centralized MARL, most state-of-art deep learning methods for network-level signal control turn to find fully or partially decentralized solutions. Pol & Oliehoek (2016) decomposed the global Q function to a linear combination of subsets of neighbor agents, which are approximated by transfer planning (Oliehoek, Whiteson, & Spaan, 2013). Combined with the max-plus coordination algorithm (Kok & Vlassis, 2005), joint actions are optimized for decision making. The main drawback of this algorithm is that the reward is designed in a global manner, which still needs fully equipped communication across the grid. Breaking down a large-scale network into disjoint sub-regions with different traffic distributions, Chu, Qu, & Wang (2016) first proposed a normalized network cutting method, then trained regional central agents to control the traffic lights of each subnetwork, based on regional observations and rewards. The critical issue with this strategy is that no communication is enhanced between manually divided regions, intersections lying at or near the borders tend to face dilemmas if two central agents issue conflicting orders. To fix this problem, T. Tan et al. (2019) applied a hierarchical MARL algorithm called 'Coder', where multiple regional agents are trained separately and a centralized global agent aggregates their value functions together to instruct and coordinate the entire network. Theoretically, this Coder algorithm does not reduce much of communication complexities in centralized control, and the system will face deadly failure if the central agent is attacked. Compared with the aforementioned works, I propose a family of decentralized MARL algorithms, where communications are only required between adjacent agents at most. There is no global or regional central agent to collect information and distribute actions among intersections. Signals, each one of which is treated as a learning agent, only exchange their local observations and immediate rewards with their neighbors, and use joint observations as states, weighted rewards as returns for TD-error propagation.

This part introduces three levels of coordination between cooperative RL signal agents: 1) No communication 2) Sharing rewards information 3) Sharing states and rewards information. These three modes will be elucidated in detail hereinafter. The intersections are connected by a communication network, I define neighbor intersections as adjacent ones (nodes) between which at least one lane (link) is shared, typically the distance from one to another is less than 1 mile. In a neighborhood pair, *m* is an upstream node of *n* (*n* is a downstream node of *m*), if $\mathcal{U}(m) \cap \Omega(n) \neq \phi$, where $\mathcal{U}(m)$ is the set of links leaving *m*, $\Omega(n)$ is the set of links entering *n*. These connections reflect not only on physical linkage, but also in that one agent's action will influence its neighbors' surrounding states directly or indirectly. Intuitively, if the upstream agent keeps giving green to vehicles going to one of its downstream intersections, then the incoming flow rate of the latter will definitely be at a higher level, and vice versa.

## 4.1 Independent Deep Q-learning

As a deeper extension to Independent Q-Learning(IQL) (M. Tan, 1993), Independent Deep Q-learning (IDQL) is proposed by (Tampuu et al., 2017) to control a pong game with two players. In the fully collaborative setting, two players (agents) train their own policies using the same philosophy in (Mnih et al., 2015), and they are rewarded an identical feedback $\rho$, "$\rho = 0$" indicating the ball is on the table, "$\rho = -1$" indicating no one catches the ball. Calvo (2018) also used the total cumulative waiting time as a global reward for all the agents when activating the IQL mode. Although these algorithms claim to implement independent/decentralized schemes, a central unit is still required to collect sensing data from agents, compute a global reward and distribute this signal to local optimizers, which in fact does not reduce any communication complexity.

The IQL for TSC in this work is fully decentralized: each one of them improves policies and makes decisions based on local observations and local rewards. A Markov game is characterized by a tuple $\left( \mathcal{S}, \{\mathcal{A}^i\}_{i \in \mathcal{N}}, \{\mathcal{O}^i\}_{i \in \mathcal{N}}, \{\mathcal{R}^i\}_{i \in \mathcal{N}}, \{\mathcal{E}^i\}_{i \in \mathcal{N}}, \mathcal{G}(\mathcal{N}, \mathcal{L}), \mathcal{T} \right)$, where

$\mathcal{N}$ is the set of intersections (nodes) in the network, $\mathcal{L}$ is the set of lanes (edges), $\mathcal{U}(i) \cap \mathcal{L} \neq \phi$ and $\Omega(i) \cap \mathcal{L} \neq \phi$, $\forall i \in \mathcal{N}$.

$\mathcal{G}(\mathcal{N}, \mathcal{L})$ is an undirected graph shaped with $\mathcal{N}$ and $\mathcal{L}$, $(i, j) \in \mathcal{G}$ means nodes *i* and *j* are connected, and this connection is bidirectional.

$\mathcal{S}$ is the state space of the entire studied traffic network, which, in this decentralized case, will be

inferred by local observations.

$\mathcal{A}^i$ is the action space of agent $i$ that $a_t^i \in \mathcal{A}^i$.

$\mathcal{O}^i$ is the observation space of agent $i$ that $o_t^i \in \mathcal{O}^i$.

$\mathcal{E}^i$ is the omission probability distribution of agent $i$ that $p(o_t^i | s_t) \sim \mathcal{E}^i$.

$\mathcal{R}^i$ is a local reward function for agent $i$, which maps the $(o_t^i, a_t^i)$ vector to a real scalar $r_t^i$, giving instant feedback if taking action $a_t^i$ under observation $o_t^i$, noted as $\mathcal{O}^i \times \mathcal{A}^i \to \mathbb{R}$.

$\mathcal{T}$ is a global state transition operator defining the probability $p(s_{t+1} | s_t)$ of entering $s_{t+1}$ given the current state $s_t$.

### 4.1.1 Markov Objects in the Context of Traffic Signal Control

How to define the Markov objects (e.g., state, action, reward) is an active topic in RL studies, and such issue is even more critical in TSC, where the number of sensors are limited. Some researchers (Gao et al., 2017) (Liang et al., 2019) (Mousavi et al., 2017) assume traffic networks are fully observable so that vehicles' positions and speeds can be obtained without omission, taking panoramic photos of intersections or rearranging equivalent information to images enables convolutional layers in DQNs to achieve satisfactory performance. Even though those methods have proven effective in labs, it is impractical to put them on roads: in states like Arkansas, New Jersey, Texas, etc (FindLaw, 2019), laws prohibit equipping intersections with cameras. To make non-biased comparison with classical and other intelligent signal control strategies, we do not add any detection advantage to the proposed algorithms, whose execution can be managed with conventional traffic detectors like dual loops and magnetic sensors.

Here are given the specific definitions of states (in this case, partial observations), actions and rewards when we train RL algorithms in a simulator. Considering the available data structure in SUMO (which will be introduced later) and the feasibility of application in engineering, we design these objects from experience.

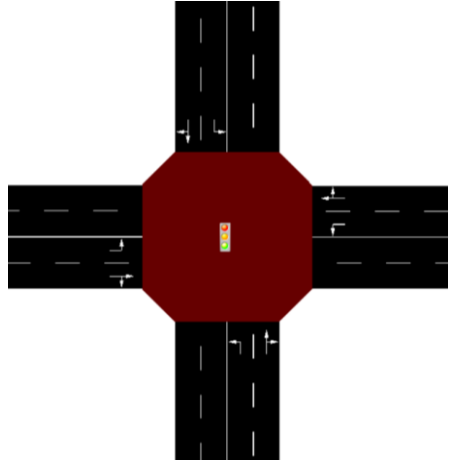

**Figure 10: Configuration of a Typical Four-way Intersection**

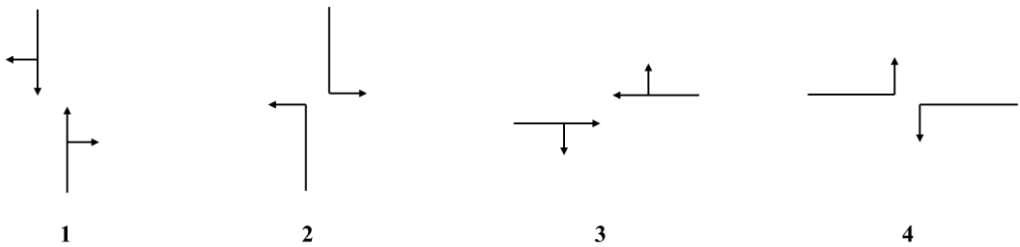

**Figure 11: Non-conflict Stage Design for a Typical Four-way Intersection**

**Action** For safety reasons, stages in traffic signal timing schedule are predefined as combinations of non-conflict phases. **Fig. 10** shows the configuration of a typical four-way intersection, there are 8 possible phases, i.e. S-N/E, S-W, N-S/W, N-E, E-W/N, E-S, W-E/S, W-N (S: South; N: North; W: West; E: East). One legal design for stages is shown in **Fig. 11**. In this case, the cardinality of the action space is 4, i.e. the number of stages. One of these stages will be activated after each time step of policy iteration, which is set as 5s in this thesis. According to the requirements of the *Caltrans Traffic Signal Operations Manual*, during the switch of two stages, an extra 3s yellow and all-red time is added to the pre-activated movements, furthermore, a maximum and minimum stage green time are mandatorily set. The minimum green time is supposed to help pedestrians cross safely, which depends on actual road conditions, a standard can be found in *Caltrans Traffic Signal Operations Manual: Appendix C*. The action trigger logic with a minimum green time is shown in **Algorithm** 8, if the elapsed time of current stage reaches the maximum green time, just switch to another one then repeat this logic.

---
**Algorithm 8** Action Trigger Logic for a Single Intersection $i$

---
**Input:** predefined stages $\mathcal{A}^i$, minimum green time $T_-$, trained control policy $\pi$
  **Initialize**:
  $\tau \leftarrow 0$
  $t \leftarrow 0$
  Activate an initial stage $a_0^i$ for 5s
  **while** Not terminated **do**
    $\tau \leftarrow \tau + 5$
    **if** $\tau >= T_-$ **then**
      Choose stage $a_{t+1}^i$ ($a_{t+1}^i \in \mathcal{A}^i$) using $\pi$
      **if** $a_{t+1}^i \;!= a_t^i$ **then**
        $\tau \leftarrow 0$
        Activate $a_{t+1}^i$ for 5s
        $t \leftarrow t + 1$
      **end if**
    **end if**
  **end while**

---

**Observation** Each agent is supposed to observe the environment locally, in this POMDP setting, the detecting range of sensors integrated at an intersection is bounded within 150 m away from its virtual geometric center. The local observation space for agent $i$ is defined as a real vector $\langle \{\mathcal{H}_l^i\}_{l \in \Omega(i)}, \{\mathcal{Q}_l^i\}_{l \in \Omega(i)}, \hbar^i, \rho^i \rangle$ with length $2|\Omega(i)|+|\mathcal{A}^i|+1$, where

$\mathcal{H}_l^i$ is the occupancy ratio of an entrance lane $l$ to intersection $i$, $l \in \Omega(i)$. Occupancy is defined as how long of the road is occupied by vehicles, for the convergence requirement in deep neural networks, $\mathcal{H}_l^i$ is normalized to [0,1], by $\mathcal{H}_l^i$ = [(5+2)•number of vehicles]/150, 5 (m) is the uniform car length, 2 (m) is the stopping gap of vehicles, 150 (m) is the sensing distance.

$\mathcal{Q}_l^i$ is the queue length ratio of entrance lane $l$ to intersection $i$, $l \in \Omega(i)$. A vehicle waiting in queues has a speed $v<0.1$m/s, this quantity can be counted by an upstream dual loop. Similar to $\mathcal{H}_l^i$, $\mathcal{Q}_l^i$ is normalized to [0,1] too.

$\hbar^i$ is a one-hot action indicator, telling which stage is currently activated. It's a 0/1 vector with length $|\mathcal{A}^i|$, only one element is 1(activated), the others are all 0. For example, if there are a total of 4 possible stages for an intersection and the second one is activated now, then $\hbar^i = [0\ 1\ 0\ 0]$.

$\rho^i$ is the elapsed time ratio of the current stage, calculated by $\rho^i = [T(a_{k+1}^i) - T(a_k^i)]/T^-$, where

$T(a_k^i)$ is the starting time of *k*-th activated stage, $T^-$ is the maximum green time.

**Reward** The local reward for each individual agent *i* is the differential cumulative waiting time between two intervals. $r_t^i = w_t^i - w_{t-1}^i$, where $w_t^i$ is the total waiting time of vehicles passing intersection *i* between time step *t*-1 and *t*. Here waiting time is defined as the time a vehicle spends when its speed is less than 0.1m/s.

### 4.1.2 Independent Deep Q Learning

`    Obviously, IDQL for a multi-agent learning task is to train each agent separately with DQN. One should notice that the original version of IDQL proposed in (Tampuu et al., 2017) still collects global information as the uniform state of each agent, and the corresponding Q network is denoted as $\hat{q}^{(i)}(s, a^i; \theta^i)$. To take a step further, agents just care about their surroundings, no extra information out of its detection distance is provided for learning. In this paper, Q networks are fed with agents' local observations, trained in the form of $\hat{q}^{(i)}(o^i, a^i; \theta^i)$. This completely decentralized method is criticized for instability and suboptimality, but is still implemented here as a benchmark.

## 4.2 Shared States Reinforcement Learning

As mentioned before, a key component of DQN is the experience replay memory. Unfortunately, the combination of experience replay with IDQL appears to be problematic because of the non-stationarity introduced by IDQL. Dynamics that generate data in the experience replay memory no longer indicate the current trajectories in which the agent is learning (J. Foerster et al., 2017). Agents learn their local optimal policies based on local observations, which in turn changes the state transition distribution of their neighbors over time.

Here I introduce another model-free decentralized cooperative MARL scheme: agents share their observations with neighbor intersections, then use both local and shared information as input states of local Q networks, with the goal to maximize expected local returns. The enriched state space (it's still partially observed, but for convenience denoted as *s*) of agent *i* at time *t* is constructed as $s_t^i = \langle o_t^i, \{o_t^j\}_{j \in \mathcal{J}(i)} \rangle$, where $\mathcal{J}(i)$ is the set of agent *i*'s neighbor nodes, $(i, j) \in \mathcal{G}$, $o_t^i$ is predefined in **4.1.1**. Similar to IDQL, shared states reinforcement learning (S2RL) follows the same independent training principle, where each agent feeds the $\{s_t^i, a_t^i, r_t^i, s_{t+1}^i, ...\}$ trajectory with prioritized experience replay into a double deep Q network $\hat{q}^{(i)}(s^i, a^i; \theta^i)$, to get an optimal Q value approximation.

### 4.3 Shared States and Rewards Reinforcement Learning

Shared states and rewards reinforcement learning (S2R2L), is an advanced version of S2RL, where agents exchange with their neighbors not only their observations but also the information of their received local rewards. This is more than adding an extra byte to the communication channel, because the RL structure is hence changed. Like a comprise between a unified global reward and completely local rewards, agents use their own instant feedback collected from local sensors and those sent from their neighbors to improve control policies. S2R2L is still a decentralized algorithm, there is no centralized or regional centralized unit controlling multiple agents, each intersection is equipped with an individual computing unit, which can learn from traffic and exert control over it.

The state space of each agent uses the same structure predefined in **4.2**, while joint rewards are designed in a weighted form $\wp_t^i = \frac{1}{n+|\mathcal{J}(i)|}(n \cdot r_t^i + \sum_{j \in \mathcal{J}(i)} r_t^j)$. The node of current interest (the main agent) is given more importance than its neighbors, whose weights are uniformly 1. $n$ is a non-negative variable here, if $n = 0$, the agent $i$ is completely 'selfless' that only considers neighbor junctions' operation, if $n \to \infty$, it jumps to the other extreme of 'selfish', just caring about itself. The selection of weight for main agents will be discussed in **Chapter 5**.

# 5 Experiments on a 4×4 Grid Network

To test the performance of the aforementioned decentralized deep MARL strategies, I apply them to the control of a 4×4 grid network, where vehicles' trips are generated with the Poisson distribution. Totally three levels of communication (i.e., IDQL, S2RL, S2R2L) are implemented to test against the benchmark performance of the Max Pressure (Varaiya, 2013) algorithm. Experiments are run in the open source microscopic traffic simulation software SUMO (Krajzewicz et al., 2012), which is a popular platform adopted by researchers in this area. The default car-following logic in SUMO is the Krauss model (Krauß, 1998), where vehicles run in collision-free protection, guaranteed to make full stops in face of potential accidents. SUMO also provides an external control interface called TraCI, allowing to retrieve values of simulated objects and to manipulate their behaviors online. As a testbed, a computational traffic control framework is built with tailored demands, which integrates advanced control algorithms and the microsimulator, bridged by TraCI (see **Fig. 12**).

Next in this section, I will first introduce the architecture of the proposed TSC framework, demonstrating how different components form a reliable platform. Second, with a 4×4 Manhattan style grid scenario, road configurations and traffic assignments are specified. Furthermore, pre-experiments are conducted to select an appropriate weight for main agents in S2RL and S2R2L. Last, the 3 decentralized MARL schemes are tested against the benchmark algorithm.

## 5.1 Architecture of Simulation Platform

Experiments conducted on this platform are able to tailor their needs according to real-world traffic operations. A network **generator** is supposed to mimic the studied road network, which can be manually drawing or with open map tools (e.g., OpenStreetMap), SUMO also provides pipelines to import network files from other simulation software like Vissim, Visum, MATsim and etc.

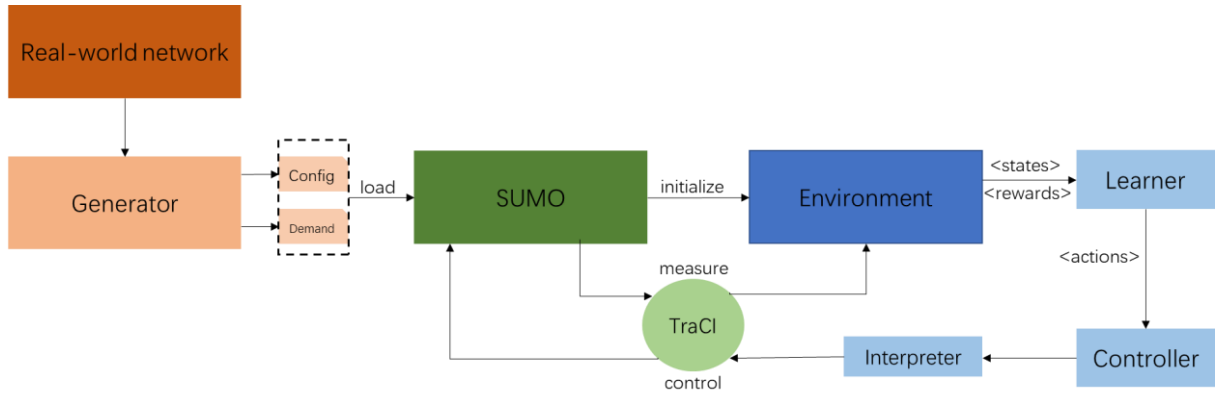

**Figure 12: Architecture of SUMO-based Traffic Signal Control Platform**

Traffic demands are assigned with predefined routes/turning ratios and flow rates, or generated with random trips. **SUMO** loads these road configuration and demand files to initialize the **environment**, functioning as the emulator of a multi-agent system. At every learning step, the environment receives measurements from SUMO simulations, including lane occupancies, queue lengths at intersections, vehicles' waiting time, currently activated phases and etc. These measurements are then sensed by **learners** as structured states (or observations) and reward signals, which are two of the three main components in the RL loop. In this context, deep neural networks play the role of learners, taking states and rewards, then outputting actions for the **controller**. Based on specified maximum and minimum green restrictions or other on-request settings, the controller is triggered to send practicable instructions to traffic lights. The intermediate step before transmitting information via TraCI is these signals should be first explained by an **interpreter**, so that lower-layer objects can be implemented.

## 5.2 Network Configuration and Traffic Assignment

**Fig. 13** illustrates the road configuration of experiments in this thesis, which is abstracted as a 4×4 Manhattan style grid network. Manhattan style refers to a grid topology where all roads only allow one-way movements and each road is divided into two lanes. On E/W (horizontal) roads, vehicles have two routing options, going straight or turning right, while on N/S (vertical) roads, permitted movements are going straight and turning left. The distance between two adjacent intersections is 150 m, a speed limit of 40 km/h is also enforced.

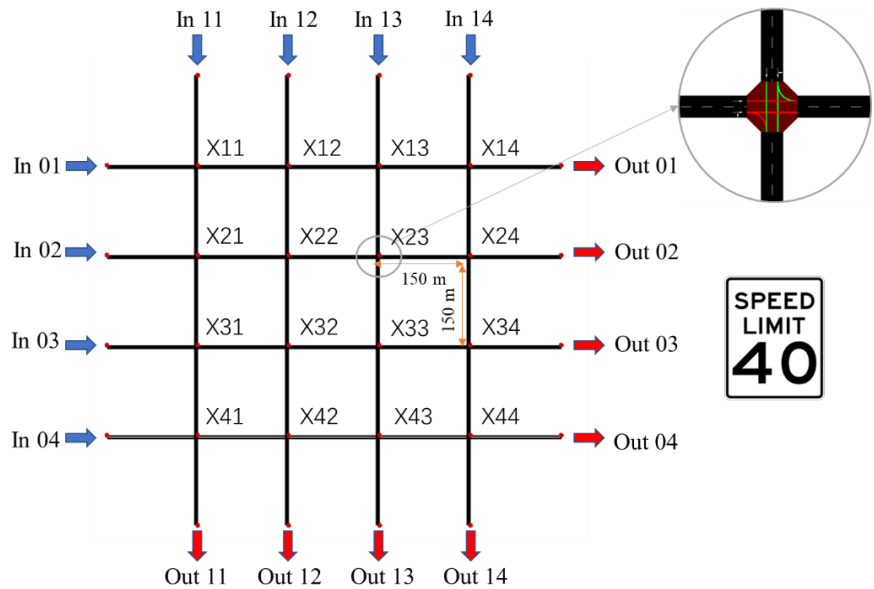

**Figure 13: A 4×4 Manhattan Style Grid Network**

Analogous to a pipe network, where the fluid is transported with specific directions and boundaries, traffic is assigned in a 'permitted Origin-Destination (OD) pair' path. We assume there exist several dummy origins and destinations on which vehicles depart and arrive. Pointed with blue arrows, dummy origins are located 150 m upstream of the leftmost and topmost nodes, denoted as *In 01 - In 04* and *In 11 - In 14* (see **Table 1**). Red arrows indicate the exits of traffic flows, which are named symmetrically as *Out 01 - Out 04* and *Out 11 - Out 14*.

Table 1 Permitted OD Pairs with Trip Generating Probabilities

|        | In 01 | In 02 | In 03 | In 04 | In 11 | In 12 | In 13 | In 14 |
|--------|-------|-------|-------|-------|-------|-------|-------|-------|
| Out 01 | p     |       |       |       | p     | p     | p     | p     |
| Out 02 | p     | p     |       |       | p     | p     | p     | p     |
| Out 03 | p     | p     | p     |       | p     | p     | p     | p     |
| Out 04 | p     | p     | p     | p     | p     | p     | p     | p     |
| Out 11 | p     | p     | p     | p     | p     |       |       |       |
| Out 12 | p     | p     | p     | p     | p     | p     |       |       |
| Out 13 | p     | p     | p     | p     | p     | p     | p     |       |
| Out 14 | p     | p     | p     | p     | p     | p     | p     | p     |

**Table 1** shows all the permitted OD pairs, alongside with a uniform trip generating probability $p \in [0,1]$. Trips between predefined ODs are released randomly with the probability of $p$ each second. On the departing time, vehicles are assigned with routes with the shortest travel time. To represent the traffic fluctuation over a day, totally 4 trip generating probabilities are alternated in a time horizon of 20000 seconds, which is the simulation time of an episode. When comparing the performance of our proposed algorithms with the benchmark, another 4 different probabilities are used in the model testing process, to evaluate the generalization abilities on unexpected traffic distributions. Details are provided in **Fig. 14**.

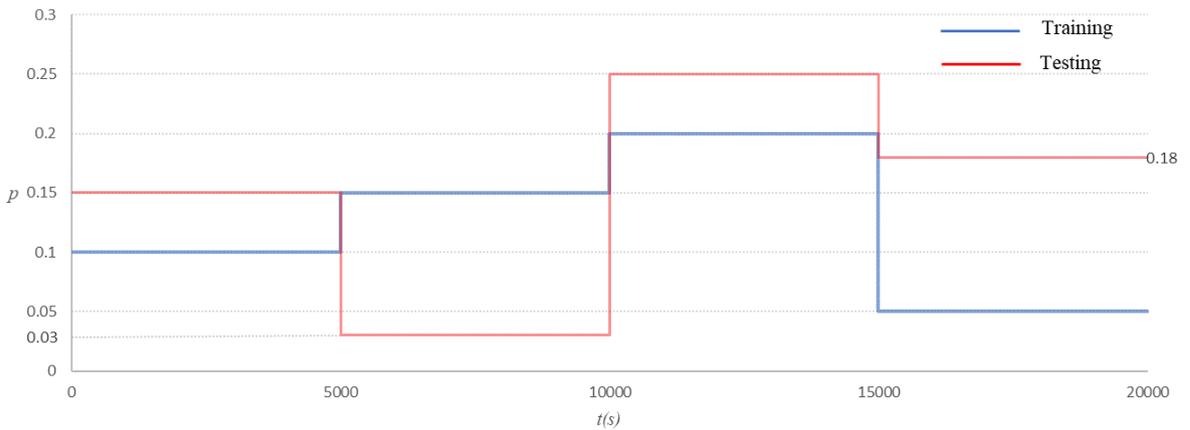

**Figure 14: Trip Generating Probabilities of Permitted OD Pairs for Training and Testing**

### 5.3 Benchmark: Max Pressure Control

The benchmark I select for testing the performance of S2R2L is the Max Pressure (MP) control

(Varaiya, 2013), whose global optimality is theoretically guaranteed. MP control was first proposed by (Ephremides, 1992) to resolve package scheduling problems in communication networks, and further proven to optimize network throughputs in signalized intersections control. It uses a concept of pressure to measure the urgency of releasing corresponding movements, and gives green to stages (combination of phases) with the highest pressure. The main reason for choosing MP is that it only requires local queue length and the information exchange between adjacent intersections, which has the same communication complexity with S2R2L. This fully decentralized framework is also feasible to realize with existing conventional loop detectors, making sure portable for practical operations.

I adapt the original MP model of (Varaiya, 2013) with the notations used in (Sun & Yin, 2018), complemented with extra explanations described below,

| | |
|---|---|
| $O_l$ | The set of all links whose starting nodes is the ending node of link $l$ |
| $I_l$ | The set of all links whose ending nodes is the starting node of link $l$ |
| $q_{l,m}(t)$ | Queue length of movement from link $l$ to link $m$ at time $t$ |
| $p_{l,m}(t)$ | Proportion of flow on link $l$ that enters link $m$ at time $t$ |
| $s_{l,m}(t)$ | Saturation flow rate on movement from $l$ to $m$ |
| $\gamma_\sigma(t)$ | Pressure associated with phase $\sigma$ at time $t$ |

As in our Manhattan style network, two lanes of an incoming branch are controlled by the same stage, links are defined as branches of intersections here, in another word, each node has 4 links, and totally 4 movements (N-S, N-E, W-E, W-S) between these links are permitted. The weight $w_{l,m}(t)$ associated with movement from link $l$ to $m$ at time $t$ is defined as the difference between $q_{l,m}(t)$ and the average queue at downstream of link $m$,

$$w_{l,m}(t) = q_{l,m}(t) - \sum_{n \in O_m} p_{m,n}(t) q_{m,n}(t) \tag{5.1}$$

Thus, the weighted pressure of a stage $\sigma$ at time slot $t$ is,

$$\gamma_\sigma(t) = \sum_{(l,m) \in \sigma} w_{l,m}(t) s_{l,m}(t) \tag{5.2}$$

where $(l, m) \in \sigma$ means movements form link $l$ to $m$ are controlled by $\sigma$, every control step, from the permitted phase set $\Sigma_i$, agent $i$ selects a stage $\sigma_i^*$ with the highest weighted pressure to activate,

$$\sigma_i^* = \arg\max_{\sigma_i \in \Sigma_i} \gamma_{\sigma_i}(t) \tag{5.3}$$

With regards to the aforementioned network, the configuration of each individual intersection is shown in **Fig. 15**, with two incoming links marked as $l_1$ and $l_2$, and outgoing links marked as $m_1$ and $m_2$. There are two available stages, $\sigma_1 = \{(l_1, m_1), (l_1, m_2)\}$ and $\sigma_2 = \{(l_2, m_1), (l_2, m_2)\}$.

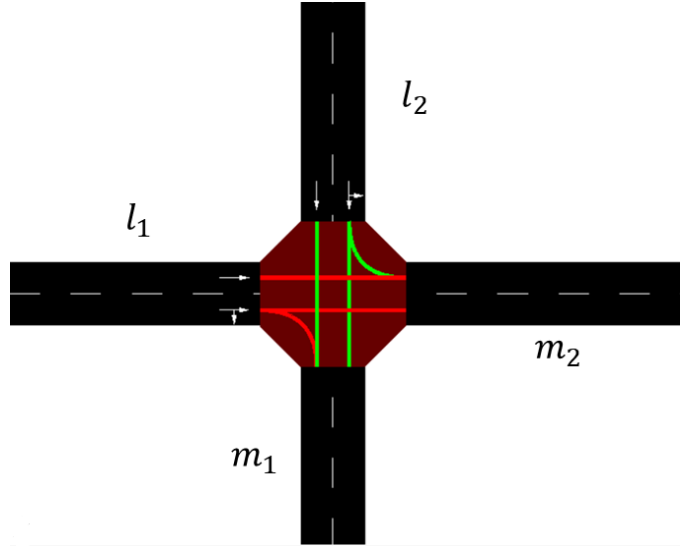

**Figure 15: Channelization of an Intersection in the 4×4 Network**

Agents make decisions every 5 seconds (one control step), activating the stage with the greater weighted pressure, with a comparison of

$$\gamma_{\sigma_1}(t) = \sum_{i=1,2} s_{l_1,m_i}(t)\left[q_{l_1,m_i}(t) - \sum_{n \in O_{m_i}} p_{m_i,n}(t) q_{m_i,n}(t)\right]$$

$$\gamma_{\sigma_2}(t) = \sum_{i=1,2} s_{l_2,m_i}(t)\left[q_{l_2,m_i}(t) - \sum_{n \in O_{m_i}} p_{m_i,n}(t) q_{m_i,n}(t)\right]$$

For homogenous intersections, saturation flow rates of straight-going movements are the same, and I make the assumption that lanes holding mixed straight/right and mixed straight/left traffic also have approximately identical saturation flow rates. Thus, the inequality relationship between $\gamma_{\sigma_1}(t)$ and $\gamma_{\sigma_2}(t)$ is directly transferred by that of $\sum_{i=1,2} q_{l_1,m_i}(t)$ and $\sum_{i=1,2} q_{l_2,m_i}(t)$. With no more need to know the queue information of upstream or downstream intersections, local agents just assign the road right to which has the longest queue length. Therefore, in this simplified setting, MP control is relaxed to the Longest Queue First algorithm (LQF, Wunderlich, Elhanany, and Urbanik 2007). To protect vulnerable road users, a minimum vehicular green time of 10 s and a maximum one of 60 s are enforced for both S2R2L and the benchmark.

## 5.4 Metrics

There are three measurements taken to evaluate the performance of proposed algorithms and the benchmark: the average vehicle delay in the network (*s/veh*), the current queue length (*veh*) and fuel consumption rate (*ml/s*) of the entire network. The delay of a vehicle is defined as the time spent below 0.1m/s since the last time it was faster than 0.1m/s, which is recorded by TraCI tagged with vehicles' ids. Every control step (5 s), the evaluator collects ids of vehicles currently in the network, then average their accumulated delay since entering the network. Queue length is intuitively the length of vehicles waiting at intersections, the aggregate value of which is also recorded every 5 seconds. The fuel consumption model implemented in SUMO is derived from PHEM and a reformulated model according to The Handbook Emission Factors for Road Transport (HBEFA v3.1), which is described in (Krajzewicz et al., 2015). Similarly, the fuel consumption rate of the entire network is calculated every 5 seconds in the unit of *ml/s*.

## 5.5 DQN Configurations

The philosophies behind IDQL, S2RL, S2R2L are different: **IDQL** considers each intersection as an isolated agent, which trains their own policies just with local observations and rewards. **S2RL** extends the state space of each agent to the combination of its own observation and that sent from neighbors, while still rewarded with local feedback. In **S2R2L**, agents not only communicate with neighbors for observation exchange, but also use a weighted form of rewards to balance 'selfish' and 'selfless'. However, one should note that all of these schemes have 16 individual agents, which represent 16 distributed controllers over 16 intersections, and each of them is supposed to follow the structure of 'n-step Double Deep Q-learning with Prioritized Experience Replay' introduced in **Algorithm 7**, the difference is how they shape their states and rewards. Here I give a well-tuned general 4-layer deep Q network model for training,

**Table 2 Configurations of Deep Q Networks**

| Layer | Number of Neurons | Feedforward | Remarks |
|---|---|---|---|
| 1 | IDQL: 11<br>S2RL/S2R2L: $(|\mathcal{J}(i)|+1)*11$ for agent $i$ | Fully connected<br>ReLU<br>Dropout ($p = 0.4$) | Feature vectors of states |
| 2 | 64 | Fully connected<br>ReLU | |
| 3 | 32 | Fully connected<br>ReLU | |
| 4 | 2 | | 2 actions |

A dropout mechanism (Srivastava et al., 2014) is embedded between the first and second layers to mitigate the overfitting problem of neural networks. When training, the linkages between these two layers have a probability of 0.4 to be removed, this "drop out" process is random, different linkages will be ignored in different batches, which prevents some neurons from achieving dominating weights. Details regarding the deep Q learning process is provided in **Table 3**

**Table 3 Parameter Setting of Deep Q Learning**

| Parameter | Value |
|---|---|
| Batch Size $b$ | 64 |
| Replay Memory Size $N$ | $10^5$ |
| TD step $n$ | 16 |
| Target Network Update Cycle $C$ | 100 |
| Exploration Rate $\varepsilon$ | $\max\left[(0.995)^t, 0.05\right]$ |
| Learning Rate $\eta$ | $10^{-4}$ |
| Reward Discount Factor $\gamma$ | 0.99 |
| Prioritized Experience Replay Exponent $\alpha$ | 1 |
| Gradient Optimization Method | Stochastic Gradient Decent (SGD) |
| Simulation Time Per Episode | 20000 s |
| Warm Time (Time to Load Vehicles) | 300 s |
| Control Step | 5 s |
| Learning Step | 16*5 s |
| Minimum Green Time | 10 s |
| Maximum Green Time | 60 s |

Instead of setting a fixed exploration rate that searches for random actions uniformly in an episode, $\varepsilon$ is assumed to adapt dynamically with a decaying form of $(0.995)^t$. Exploring on the action space during early stages of training is given preference, but it must make a concession when the policy approaches convergence, until touches the bottom value of 0.05. To initialize the environment, a warm-up time of 300 s allows SUMO load vehicles to the network, before they are sensed and controlled by DQN agents. During the warm-up time, traffic lights are scheduled by the MP algorithm.

### 5.6 Results

#### 5.6.1 Pre-Experiments for Rewards' Weight Decision of Main Agents

In S2R2L, agent $i$ uses a weighted reward $\wp_t^i = \frac{1}{n+|\mathcal{J}(i)|}(n \cdot r_t^i + \sum_{j \in \mathcal{J}(i)} r_t^j)$ to loop the iteration process of local Q function approximation. The critical problem is how we decide the reward weight of itself (we call the main agent here), as this is the key component to balance "selfish" and "selfless". To find a relatively optimal value for this hyperparameter, we recommend 9 candidates as 0, 0.5, 1, 2, 3, 5, 10, 100, 1000. First, we tune 9 models separately with the training demand (see **Fig. 14**), then apply fine-tuned models (trained 50 episodes) to the testing assignment, evaluated by metrics stated in **5.4**.

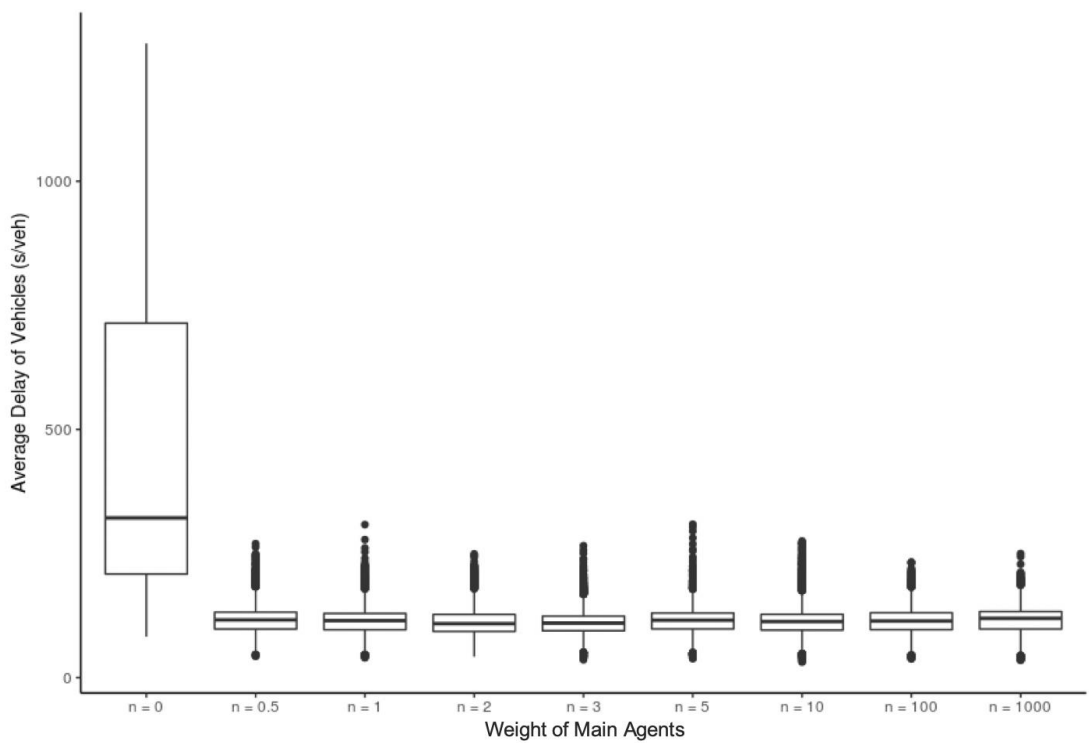

**(a1) Average Vehicle Delay with Different Weights**

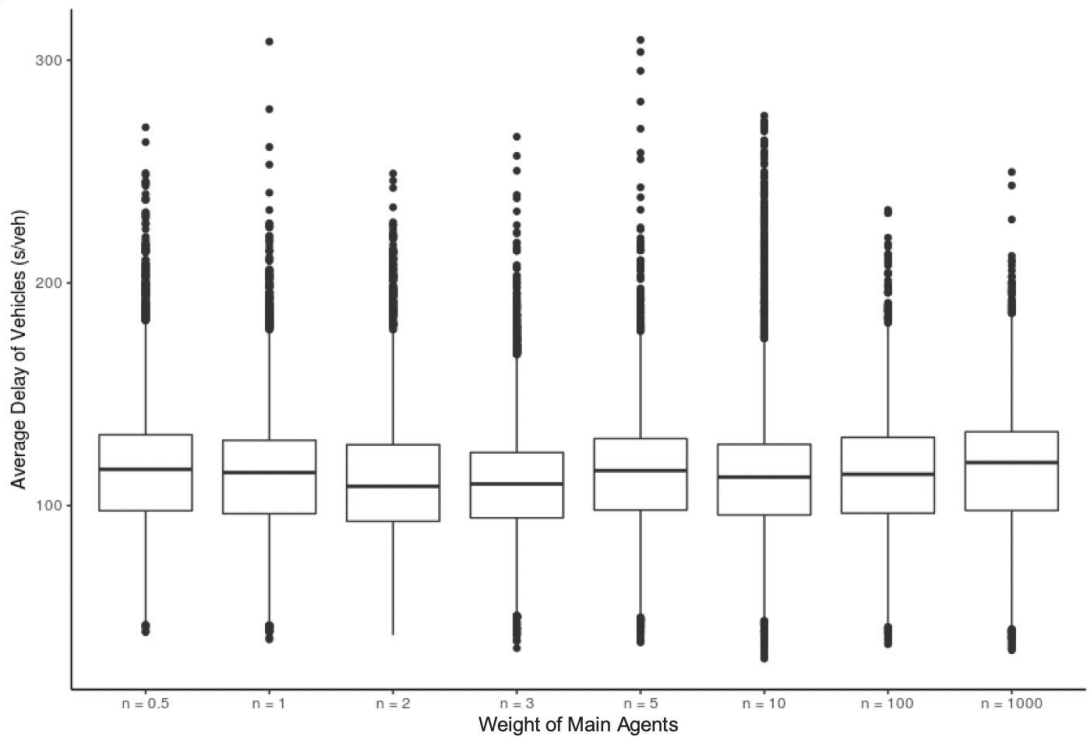

**(a2) Average Vehicle Delay with Different Weights (Except n = 0)**

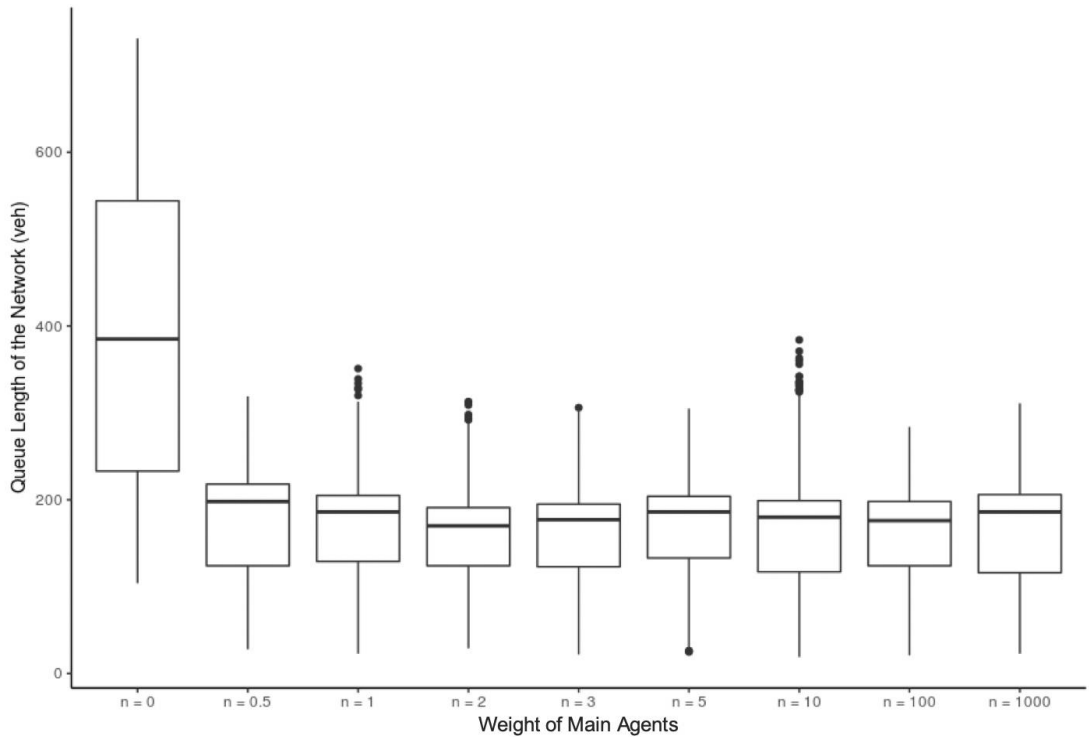

**(b) Number of Queued Vehicles of the Network with Different Weights**

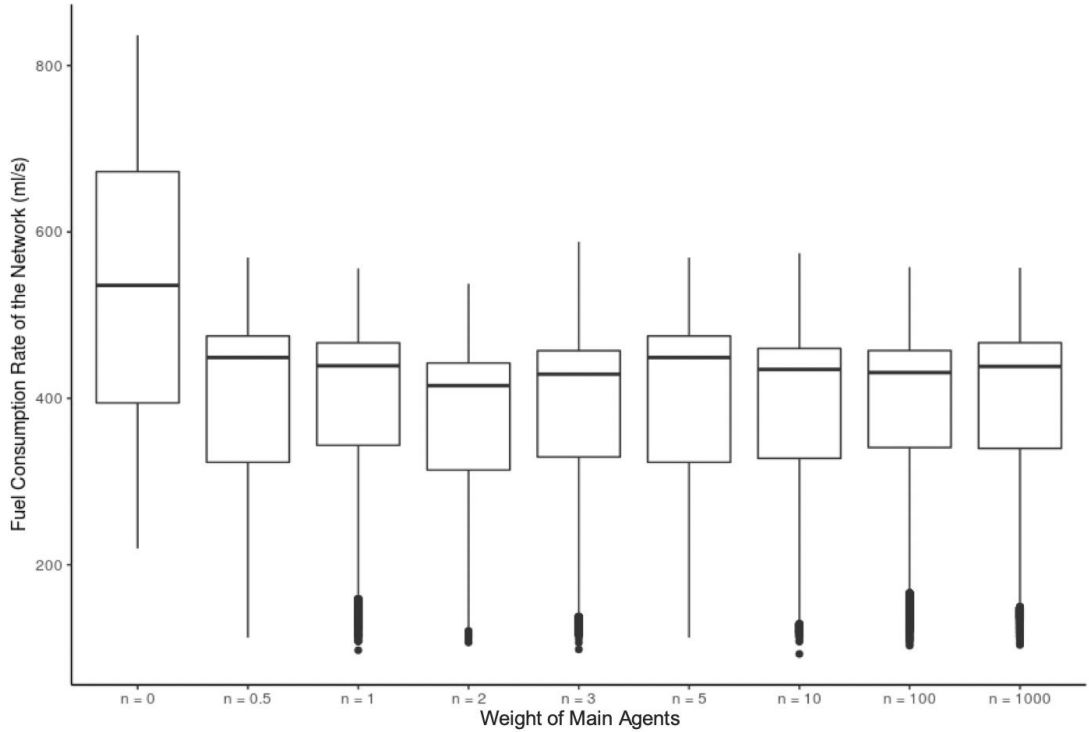

**(c) Fuel Consumption Rate of the Entire Network with Different Weights**

**Figure 16: Performance of S2R2L with Different Reward Weights of Main Agents**

**Fig. 16** suggests that S2R2L with main agents' reward weight of 2 has the best performance. On the three metrics, the median value of '$n = 2$' is the lowest among 9 candidates, and the quantile boundaries of which are more compact than those of the others, indicating this weight makes the traffic smoother. Meanwhile, '$n = 0$' is definitely the worst choice, which induces longer queue length, more vehicle delay and more fuel consumption of the network. This phenomenon is explainable since the agents just care about how well their neighbors operate, totally forgetting the situation of themselves. When $n = 1000$, S2R2L is approximately equivalent to S2RL, whose performance is also not competitive. Based on the results of the pre-experiment, $n = 2$ is taken as the reward weight of main agents in S2R2L. Next in this chapter, the comparison is conducted between MP, IDQL, S2RL, S2R2L ($n = 2$).

### 5.6.2 Training Performance

Since the reward definitions for IDQL, S2RL, S2R2L are different, it is hard to plot the reward evolution traces over episodes, which is a common standard to test the convergence performance of RL algorithms. Instead, I collect the values of the aforementioned 3 metrics of these 3 MARL schemes every control step (5 s), then average them over one episode to see how the models advance to the optimum.

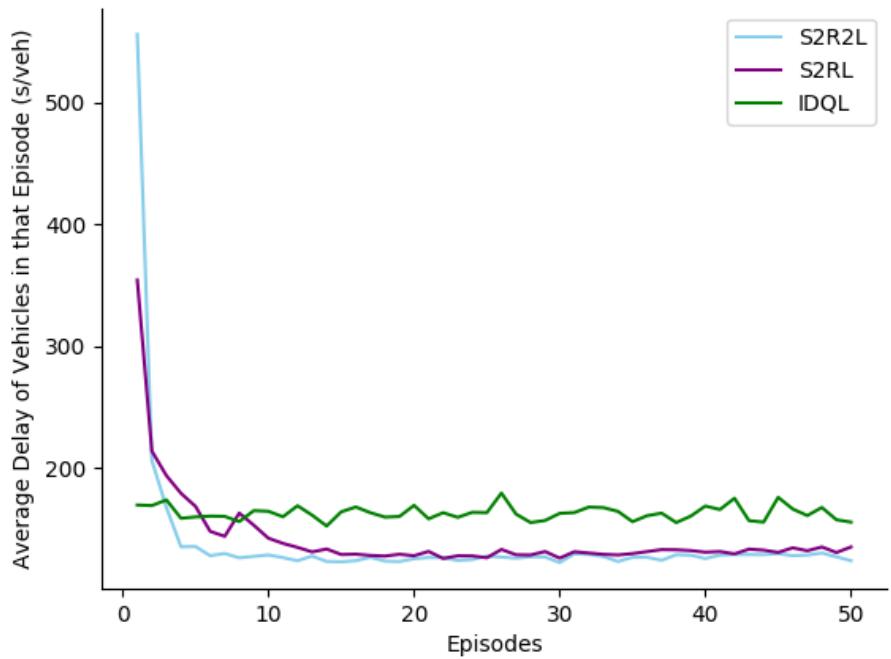

**(a) Average Vehicle Delay of Each Episode**

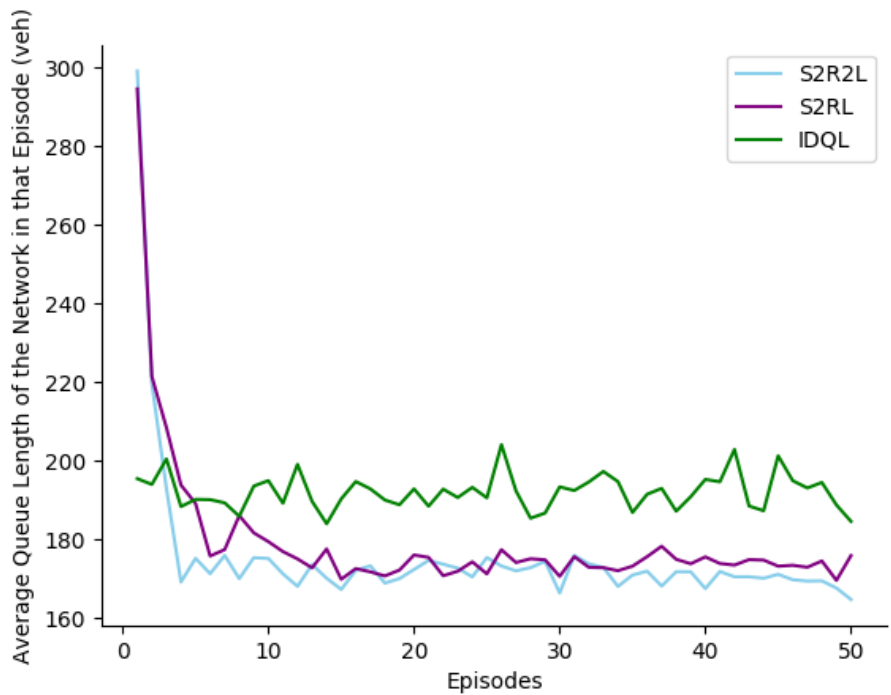

**(b) Average Number of Queued Vehicles of the Network of Each Episode**

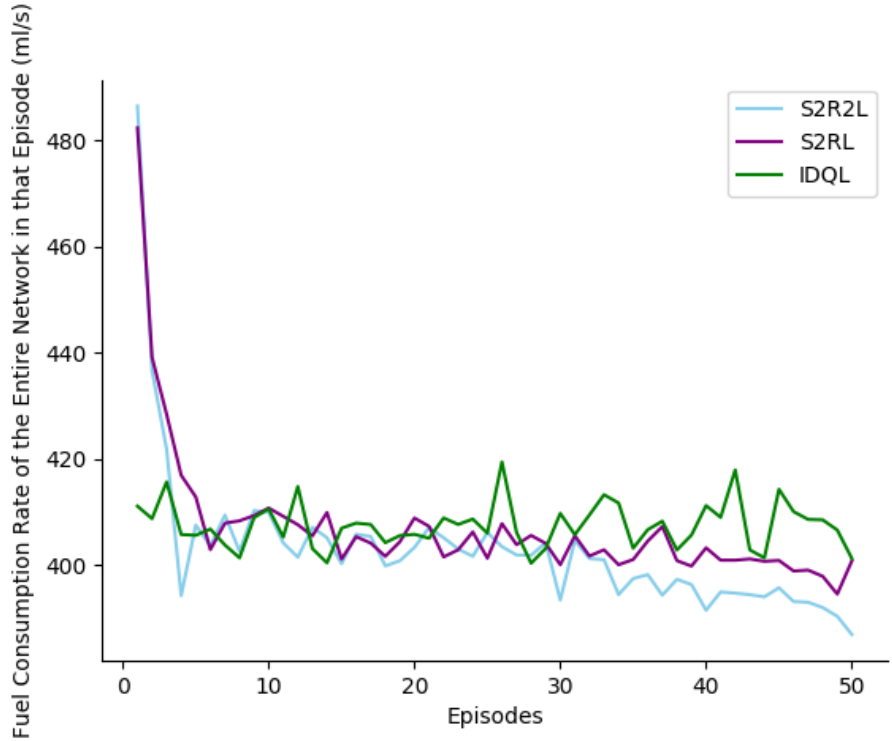

**(c) Average Fuel Consumption Rate of the Entire Network of Each Episode**

**Figure 17: Training Performance of 3 MARL Algorithms**

**Fig. 17** illustrates S2R2L has the lowest values under all the 3 criteria, which means it helps this signalized network achieve the best promise in reducing traffic delay, waiting vehicle numbers at intersections and fuel use. When the number of training episodes goes up, S2RL is guaranteed to approach S2R2L in vehicle delay mitigation, while a small gap still exists when these two shared-rewards MARL schemes are examined on reducing network-level queue length and fuel usage rate. S2RL and S2R2L both show convergence trends as more episodes are trained, and the latter has a faster convergence speed. On the opposite, IDQL reports the worst performance, not only in the measurements of 3 metrics, but also for its poor convergence. Lacking communication with other agents for information exchange, the partially observable state spaces of agents in IDQL is unstable, as a consequence, their learned policies quickly expire since the transition distribution of observations have changed. On this point, the training results support the criticism of IDQL in previous studies.

## 5.6.3 Testing Performance

Testing experiments are conduced based on a different travel demand distribution from the one used in training (See **Fig. 14**). The total simulation time is 20000 s, representing the same period in real-world. Using Max Pressure algorithm as the benchmark, simulation results are displayed in **Fig. 18**.

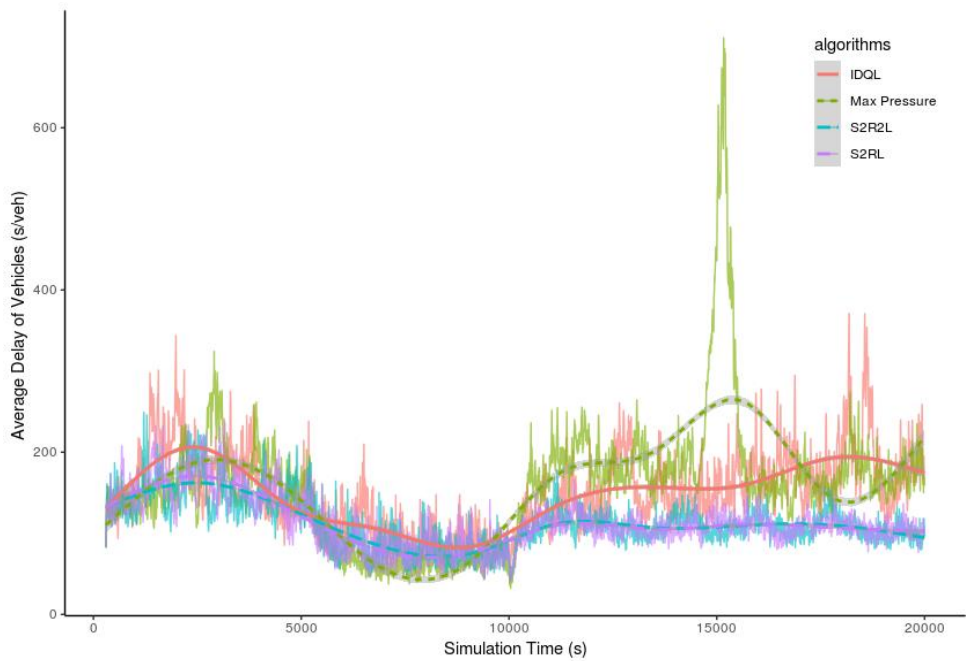

**(a) Average Vehicle Delay**

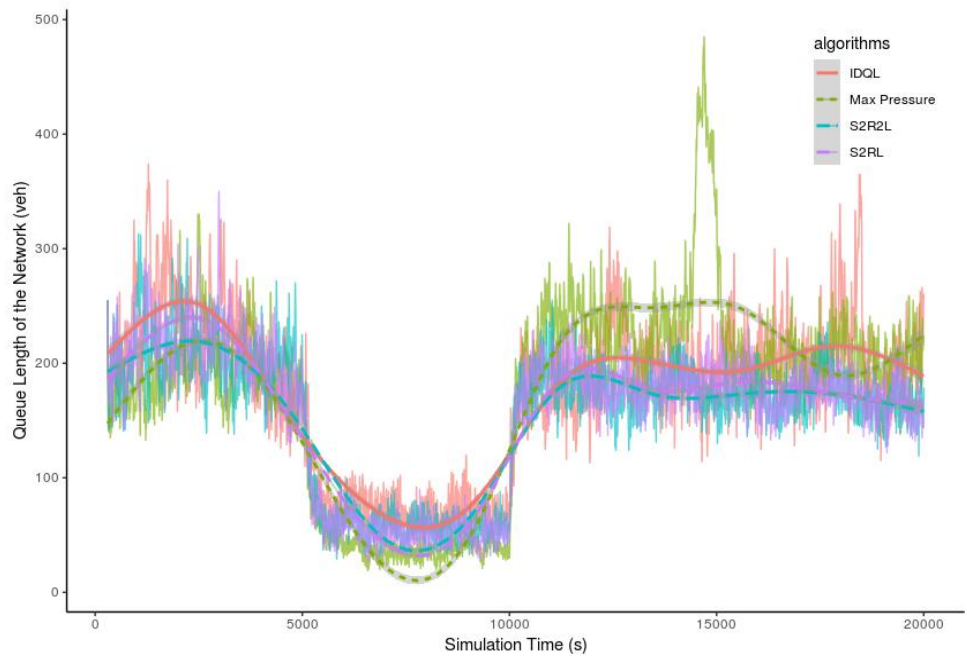

**(b) Average Number of Queued Vehicles of the Network**

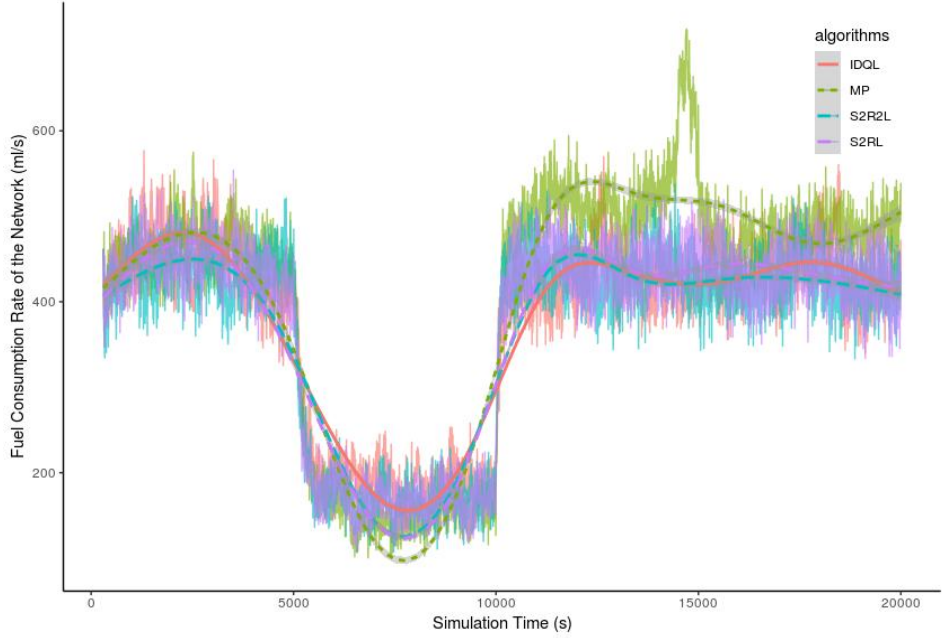

**(c) Fuel Consumption Rate of the Entire Network**

**Figure 18: Testing Performance of 3 MARL Algorithms and Max Pressure**

I roughly divide these 4 traffic demands (each lasts for 5000 s) into 3 categories: Low ($p = 0.03$), Medium ($p = 0.15$ and $p = 0.18$) and High ($p = 0.25$). Curves in **Fig. 18** well reflect the corresponding trend of traffic fluctuation. With few vehicles arriving (5000 s – 10000 s), MP guides traffic signals in a more deterministic way, thus reducing average delay, queue length and fuel consumption rate to the most. Intuitively, when the demand is low, all of these 4 algorithms can serve as a reliable control strategy, functioning better than under the other 3 demands. That is to say, the expected global state value $V(s)$ associated with these states is higher than those of a congested network. During the first medium demand period (0 – 5000 s), S2R2L achieves comparable performance with MP in queue reduction and even higher scores in the other two metrics. This advantage becomes overwhelming when the trip generation rate jumps from 0.03 to 0.25, where MP fails to accommodate surging traffic just with the longest-queue-first strategy. S2RL and S2R2L still maintain their performance superiority when the demand level goes back to medium at 15000 s, when all the three indexes of MP control pulsed to a peak. During the later stage of 'high demand', more and more vehicles get stuck in traffic in the MP-controlled network, queues spilling back, idling cars consuming more gas. Not until the network demand decreases does this dilemma get eased, but S2RL and S2R2L has shown that they can deal with a larger range of traffic demands more effectively than MP and IDQL.

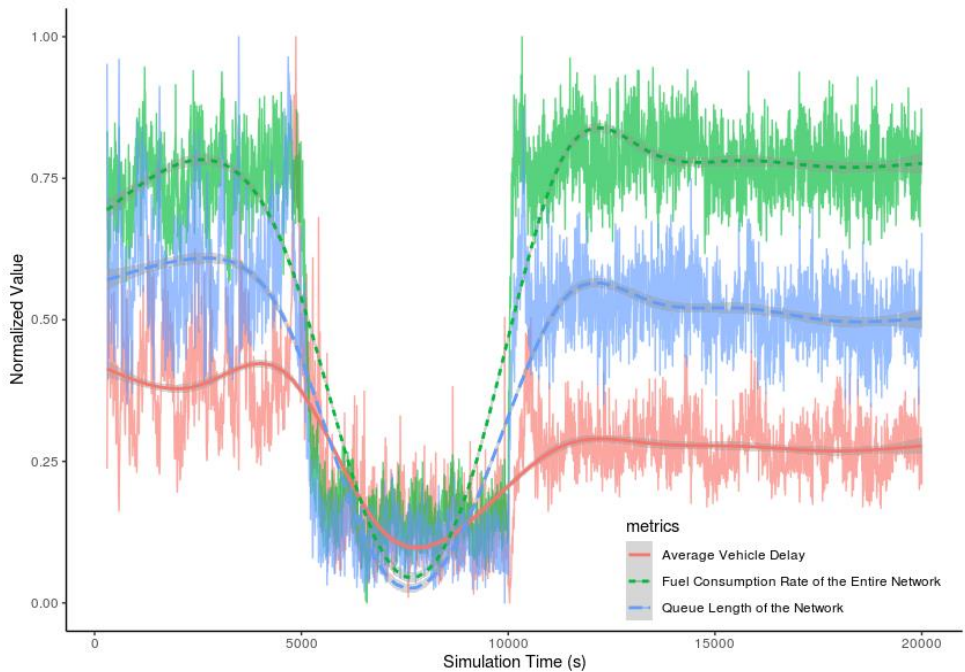

(a) Normalized Values of Metrics of S2R2L

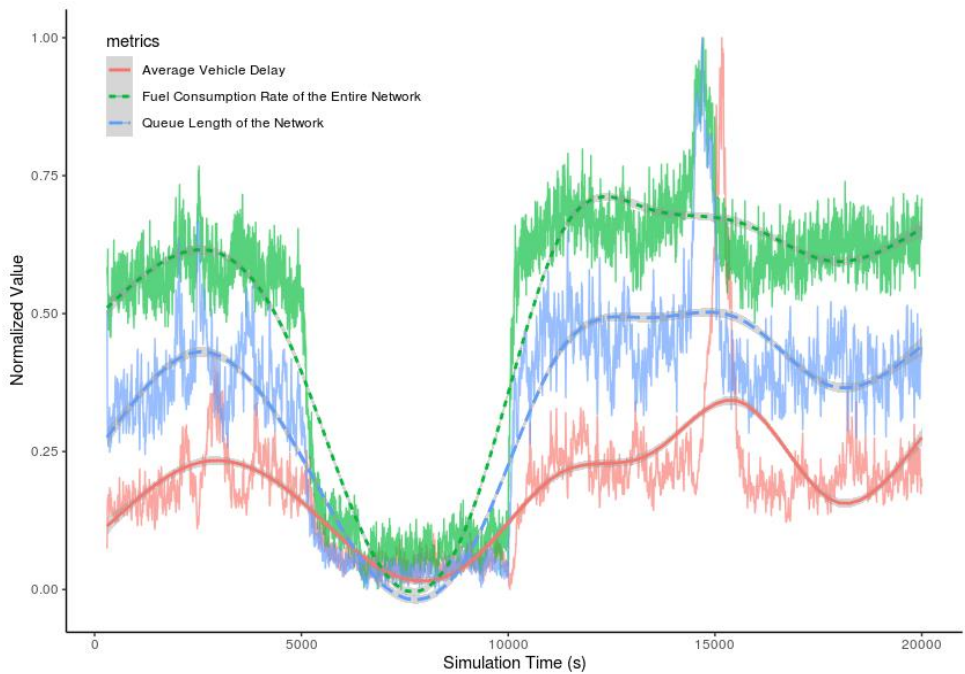

(b) Normalized Values of Metrics of Max Pressure

Figure 19: Normalized Training Performance of S2R2L and Max Pressure

I further normalize the values of three metrics to [0,1] for S2R2L and MP, plotting the temporal trend in **Fig. 19** to see how these indices are related. It is obvious that for both algorithms, the fuel consumption rates of the entire network are most sensitive to the fluctuation of traffic demand, while the index of average vehicle delay has the minimum range. We can also tell the total number of queued vehicles and the fuel consumption rate of the entire network change synchronously, more vehicles waiting in the queue, more fuel is consumed. However, there exists a time-delay between the average vehicle delay and the other two metrics. It is more evident when we look at the normalized curve of MP that the peak of delay shifted right to that of the other metrics. One possible explanation for this phenomenon is that the pulsed traffic has a time-lag influence on vehicles' delay, since in this thesis, 'delay' is calculated as the cumulative time of vehicles spend when their speeds are lower than 0.1m/s.

Taking a more microscopic view, we look at how well these schemes work on 16 individual intersections. **Tables 4** and **5** record the statistics of average vehicle delay and average queue length at each intersection. The results show that IDQL, S2RL and S2R2L all surpass the benchmark MP control under these two criteria, with improvements of 12.78%, 33.72%, 34.55% in delay reduction and 1.75%, 8.98%, 10.91% in queue dissipation, correspondingly. It is worth noting that S2RL and S2R2L does not outperform MP control at all intersections despite that their overall performance is superior. For example, at intersections X23 and X24 S2RL and S2R2L produce longer delay and at X21 they produce longer queues than MP control.

Table 4 Intersection-level Average Vehicle Delay

|  | Average Delay (s/veh) | | | | % Improvement | | |
|---|---|---|---|---|---|---|---|
|  | MP | IDQL | S2RL | S2R2L | IDQL vs MP | S2RL vs MP | S2R2L vs MP |
| X11 | 91.9 | 54.1 | 46.3 | 46.7 | 41.13 | 49.62 | 49.18 |
| X12 | 142.5 | 84.0 | 66.8 | 73.8 | 41.05 | 53.12 | 48.21 |
| X13 | 207.7 | 279.5 | 200.6 | 199.8 | -34.57 | 3.42 | 3.80 |
| X14 | 220.6 | 298.6 | 213.1 | 216.1 | -35.36 | 3.40 | 2.04 |
| X21 | 209.2 | 115.3 | 97.7 | 98.9 | 44.89 | 53.30 | 52.72 |
| X22 | 202.6 | 305.5 | 180.4 | 171.8 | -50.79 | 10.96 | 15.20 |
| X23 | 219.4 | 341.9 | 227.2 | 220.6 | -55.83 | -3.56 | -0.55 |
| X24 | 248.8 | 346.9 | 264.6 | 269.1 | -39.43 | -6.35 | -8.16 |
| X31 | 180.2 | 105.8 | 80.6 | 73.8 | 41.29 | 55.27 | 59.05 |
| X32 | 218.6 | 100.5 | 82.6 | 77.2 | 54.03 | 62.21 | 64.68 |
| X33 | 151.3 | 89.5 | 78.1 | 73.4 | 40.85 | 48.38 | 51.49 |
| X34 | 182.2 | 168.1 | 126.4 | 136.1 | 7.74 | 30.63 | 25.30 |
| X41 | 190.2 | 211.7 | 153.4 | 154.6 | -11.3 | 19.35 | 18.72 |
| X42 | 190.9 | 246.1 | 155.7 | 156.1 | -28.92 | 18.44 | 18.23 |
| X43 | 198.6 | 111.1 | 98.7 | 87.9 | 44.06 | 50.30 | 55.74 |
| X44 | 225.4 | 239.0 | 166.8 | 171.6 | -6.03 | 26.00 | 23.87 |
| Network | 181.5 | 158.3 | 120.3 | 118.8 | 12.78 | 33.72 | 34.55 |

Table 5 Intersection-level Average Queue Length

|  | Average Queue Length (veh) | | | | % Improvement | | |
|---|---|---|---|---|---|---|---|
|  | MP | IDQL | S2RL | S2R2L | IDQL vs MP | S2RL vs MP | S2R2L vs MP |
| X11 | 20.0 | 17.8 | 16.8 | 16.6 | 11.00 | 16.00 | 17.00 |
| X12 | 14.1 | 14.0 | 12.8 | 12.8 | 0.71 | 9.22 | 9.22 |
| X13 | 6.5 | 5.7 | 6.4 | 5.6 | 12.31 | 1.54 | 13.85 |
| X14 | 6.9 | 4.9 | 5.6 | 5.6 | 28.99 | 18.84 | 18.84 |
| X21 | 11.6 | 15.5 | 13.1 | 12.5 | -33.62 | -12.93 | -7.76 |
| X22 | 9.3 | 9.5 | 9.0 | 8.5 | -2.15 | 3.23 | 8.60 |
| X23 | 7.9 | 6.7 | 7.5 | 7.3 | 15.19 | 5.06 | 7.59 |
| X24 | 6.0 | 5.0 | 4.9 | 4.4 | 16.67 | 18.33 | 26.67 |
| X31 | 12.1 | 14.5 | 12.1 | 10.8 | -19.83 | 0 | 10.74 |
| X32 | 11.2 | 12.7 | 10.7 | 11.1 | -13.39 | 4.46 | 0.89 |
| X33 | 15.8 | 15.8 | 14.0 | 13.0 | 0 | 11.39 | 17.72 |
| X34 | 11.2 | 9.4 | 8.4 | 9.0 | 16.07 | 25.00 | 19.64 |
| X41 | 8.3 | 9.0 | 7.0 | 7.7 | -8.43 | 15.66 | 7.23 |
| X42 | 7.9 | 7.4 | 6.9 | 8.0 | 6.33 | 12.66 | -1.27 |
| X43 | 14.4 | 11.8 | 13.1 | 13.0 | 18.06 | 9.03 | 9.72 |
| X44 | 8.2 | 8.8 | 7.7 | 7.0 | -7.32 | 6.10 | 14.63 |
| Network | 171.4 | 168.4 | 156.0 | 152.7 | 1.75 | 8.98 | 10.91 |

### 5.6.4 Sensitivity Analysis on the Maximum Green Time

To test the generalization ability of S2R2L and to figure out whether the predefined maximum green timing (60 s) is reasonable, I applied a sensitivity analysis in this section by assuming that the maximum green time is required to be set as other values in practical operation. The S2R2L model is fixed after trained with the settings described in **Table 3**, and tested under the restriction of 4 different maximum green times (30s/40s/50s/60s). For comparison, the Max Pressure control is also implemented with corresponding restrictions. The testing traffic demand is plotted in **Fig. 14**, aggregate results are shown in **Table 6**.

**Table 6 Testing Performance of S2R2L and Max Pressure with Different Maximum Green Time**

|  | S2R2L | | | | Max Pressure | | | |
| --- | --- | --- | --- | --- | --- | --- | --- | --- |
| Max Green (s) / Metrics | 30 | 40 | 50 | 60 | 30 | 40 | 50 | 60 |
| Average Vehicle Delay (s/veh) | 184.6 | 135.9 | 120.3 | 118.8 | 190.2 | 148.3 | 152.7 | 181.5 |
| Averge Queue Length of the Entire Network (veh) | 175.6 | 159.3 | 154.1 | 152.7 | 177.3 | 161.5 | 163.2 | 171.4 |
| Average Fuel Consumption Rate of the Entire Network (ml/s) | 410.2 | 391.5 | 379.9 | 374.9 | 415.6 | 395.5 | 396.1 | 403.7 |

As shown in **Table 6**, S2R2L still has a comparable performance with the original setting though the maximum green time is lowered to 50s, while the quantities of the three metrics quickly increase as the max green keeps going down. When the max green is on the lower bound of 30s, S2R2L gets the poorest scores, which are even worse than those of MP with 60s max green. In the contrast, when the forced signal switching thresholds are adjusted to 50s and 40s, MP has to terminate the green for phases with continuous high demand earlier. This helps MP overcome the 'myopia' caused by just thinking of the current situation. With the max green fixed as 40s, MP gets the best performance, however, it fails to control the traffic well when the threshold reaches 30s. It is worthy that S2R2L still outperforms MP with the same max green restriction.

**5.6.5 Look into the Black Box**

One inherent defect of deep learning-based algorithms is that these models depend heavily on the error propagation of neural networks, which functions more like a black box, making interpretation of parameters difficult. Such weakness is amplified when the applied scenario is sensitive to disturbance, take the TSC task for example, a mistrained MARL controller tends to make the traffic network unstable, even cause fatal accidents. Building explainable deep learning models is recognized as the next epoch-making effort in the AI fields (Gilpin et al., 2019), thus I give some insights into how S2R2L works as a reliable signal controller.

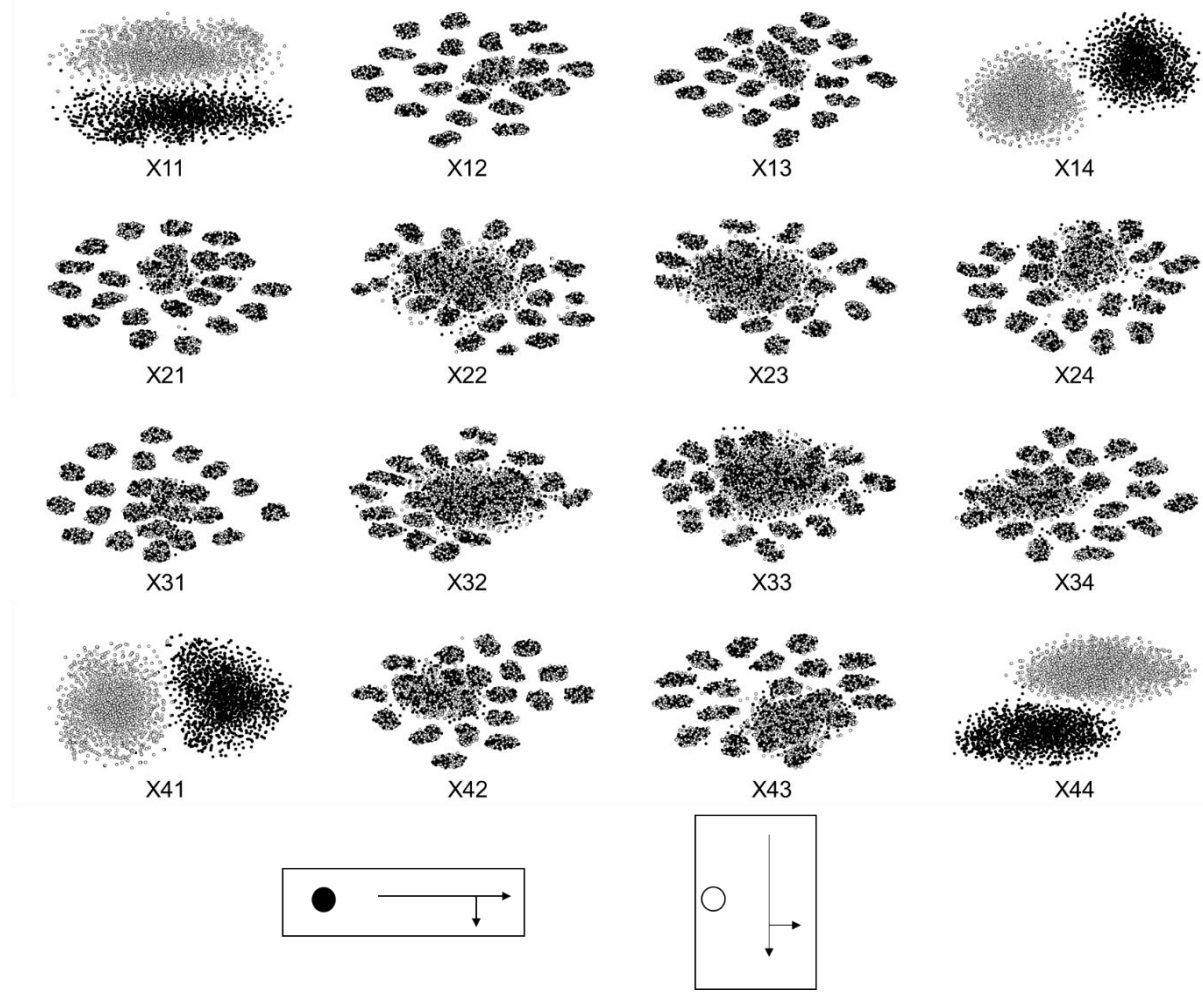

**Figure 20: Two-dimensional t-SNE Embedding of the First Layer of S2R2L**

In the context of practical TSC operation, the controller follows a 3-step procedure of train-store-apply: models are first trained in a simulator, then store fine-tuned parameters to the memory, during real-world control, the fixed model does a multiple-input single-output (MISO) mapping, taking sensed traffic features in, sending out a recommended action to the actuator. Here I use a technique called t-distributed stochastic neighbor embedding (t-SNE) (Van Der Maaten & Hinton, 2008) to analyze this mapping. t-SNE is a powerful tool for high-dimensional data visualization, it tends to gather data points with similar characteristics (measured by a standard described in their paper) and to minimize the Kullback–Leibler divergence between two distributions. In our case, each data point represents the local states of an agent (a vector with cardinality of $(|\mathcal{J}(i)|+1)*11$), they are clustered to a 2-dimensional space by mutual similarity, embedded with corresponding mapped actions (in case of skipping for minimum/maximum green, I still tag the actions output from DQNs).

**Fig. 20** shows the t-SNE plots of the 16 agents for S2R2L, arranged by their positions in the grid network. Interestingly, these 16 graphs make a symmetric appearance, states of the 4 corner intersections are explicitly divided into 2 categories, and the corresponding agents have a clear brain to make deterministic decisions. Though the state-action correlations of the other agents are vague, we can still find some similarity of state characteristic distribution between topologically symmetric nodes. The main clusters of inner nodes (X22, X23, X32, X33) are bigger than those of nodes on edge, which indicates the traffic features at these intersections are more complicated to figure out than the others. Looking more closely, the t-SNE plots of these 4 nodes can be roughly seen as copies rotationally transformed from either one of them, complementary experiments are required to verify this finding.

## 5.7 Conclusion

To test the performance of proposed three MARL algorithms, a SUMO-based simulation platform is developed to mimic the traffic evolution of real world. Fed with random trips between permitted OD pairs, a 4×4 Manhattan-style grid network is set up as the testbed, totally two different vehicle arrival rates are generated. First, according to the results of pre-experiments, *n* = 2 is determined as the reward weight for main agents out of 9 candidates. Second, S2R2L shows a quicker convergence speed and better convergent performance than IDQL and S2RL in the training process. Third, assigned with a trip generation rate different from that used for training, three MARL schemes all show good generalization abilities. The testing results of them surpass the benchmark Max Pressure algorithm, under the criteria of average vehicle delay, the total number of queued vehicles and the fuel consumption rate of the entire network. Notably, the network-level queue length and fuel consumption rate change synchronously, while there exists a time-lag between the average vehicle delay and the other two metrics. In total, S2R2L has the best testing performance of reducing up to 34.55% traffic delay and up to 10.91% queue length compared with MP. To test the generalization ability of S2R2L, a sensitivity analysis on the maximum green time is conducted, the results show that S2R2L still has satisfactory performance when the max green is lowered to 50s, and surpasses MP with the same max green restriction. Furthermore, I apply t-SNE to visualize the state distributions of individual intersections, embedded with actions recommended by S2R2L. Agents on the four corners always take deterministic actions since their states are divided more explicitly, some symmetric features are also found in t-SNE plots of the 4 inner nodes.

# 6 MARL TSC Beyond Simulation

Our previous MARL based TSC training and tests relied on a simulated environment (SUMO), because training a MARL algorithm requires tons of samples, which is made possible only through simulation given the time frame for this work. Intuitively, we cannot train the models from start in a real traffic signal control environment due to safety concerns. The exploration mechanism in RL makes the situation even worse, if signals are allowed to search for random actions, the traffic will highly likely end up to be inchaos. Therefore, the common case in advanced TSC research is to first develop an efficient and robust model on well-calibrated simulation platforms, then apply that reliable model to practical operation. The challenge with this philosophy is that even the best available simulators are hard to reflect all the important dynamics of reality. "All models are wrong, but some are useful." However, in fact, useful models trained in emulators fail to generalize to complex situations in reallity. This discrepancy between simulation and real world is called the "reality gap", which exists but is rarely addressed in previous studies.

This "reality gap" or "optimality gap" has been attracting more and more attention from the community of robotic control. Jakobi, Husbands, & Harvey (1995) developed a method to add artificial independent identical distributed (i.i.d) noise to simulators' sensors and actuators so that the learner would shift its focus from too detailed features, thus to avoid the simulation optimization bias. Domain randomization (Muratore, Gienger, & Peters, 2019) is another state-of-art remedy from a Bayesian point of view, it slightly randomizes the simulator's parameters to intentionally put models under environmental uncertainties. As a result, when these models encounter systems with similar distributions but different parameters, they show a more robust performance than those without domain randomization. More discussion are available in (Shrivastava et al., 2017), (Bousmalis et al.,

2016), and (Ganin et al., 2017).

The Markov object $\mathcal{M} = \{\mathcal{S}, \mathcal{A}, \mathcal{O}, \mathcal{R}, \mathcal{E}, \mathcal{T}\}$ distinguishes a POMDP, thus the environmental dynamics of a reinforcement learning algorithm. We make the assumption here that both the simulated and real-world systems use the same detection technology (e.g., dual loop detectors) to sense the environment and the same expert experience to map the state $s$ to feature vector $X(s)$. Moreover, with an identical reward definition, which is also dependent on traffic sensing, the "reality gap" in RL problems shrinks to the gap of state transition distributions between simulation and reality. In this " outlook of future research" section, I give a conceptual policy-based reinforcement learning modification to close this gap, making RL a more feasible and safer scheme for TSC.

Policy-based methods are on the other hand of two main branches in RL, which, in contrast to value-based ones, directly map states to actions, without Q functions to make a maximum search. The goal of deep policy-based RL is to find the optimal parameters $\theta^*$ of a deep policy network $\pi(a|s;\theta)$ that maximize the expected return of $\bar{R}_\theta$, also written as $J(\theta) = E_{\tau \sim p_\theta(\tau)}[\sum_t r(s_t, a_t)]$. Policy Gradient (Williams, 1992) is an essential technique to update parameters in the direction of $\nabla_\theta J(\theta)$. With the intention to eliminate the sampling variance, we also introduce the concept of parallel computing that at the beginning of each episode, the configurations of the simulation system are duplicated to several copies, where multiple RL agents interact with the environment and collect samples simultaneously. For an N-copy parallel episodic setting, the gradient of $J(\theta)$ is formulated as,

$$\begin{aligned}
\nabla_\theta J(\theta) &= \nabla_\theta [\sum_\tau R(\tau) p_\theta(\tau)] = \sum_\tau R(\tau) p_\theta(\tau) \frac{\nabla_\theta p_\theta(\tau)}{p_\theta(\tau)} \\
&= \sum_\tau R(\tau) p_\theta(\tau) \nabla_\theta \log p_\theta(\tau) \\
&= E_{\tau \sim p_\theta(\tau)} [R(\tau) \nabla_\theta \log p_\theta(\tau)] \\
&\approx \frac{1}{N} \sum_{n=1}^{N} R(\tau^n) \nabla_\theta \log p_\theta(\tau^n) \\
&= \frac{1}{N} \sum_{n=1}^{N} \sum_{t=1}^{T_n} R(\tau^n) \nabla_\theta \log p_\theta(a_t^n \mid s_t^n) \\
&\approx \frac{1}{N} \sum_{n=1}^{N} \left( \sum_{t=1}^{T_n} \left( \sum_{t=1}^{T_n} r_t^n \right) \nabla_\theta \log p_\theta(a_t^n \mid s_t^n) \right)
\end{aligned} \quad (6.1)$$

Where $T_n$ is the total simulation time of copy $n$. Typically, there are two improvements to further reduce variance: considering the causality and adding a baseline. The causality of sequences is that the action taken at time $t$ only influences the trajectories behind it, therefore, instead of taking the accumulated rewards of an entire episode as the uniform weight, the learner just care about how much an action contributes to later rewards,

$$\nabla_\theta J(\theta) \approx \frac{1}{N} \sum_{n=1}^{N} \left( \sum_{t=1}^{T_n} \left( \sum_{t'=t}^{T_n} r_{t'}^n \right) \nabla_\theta \log p_\theta(a_t^n \mid s_t^n) \right) \quad (6.2)$$

There is another case that we add an unbiased baseline $b = \frac{1}{N} \sum_{n=1}^{N} R(\tau^n)$ so that the cumulative reward terms are not always positive, helping reduce high variance. The subtraction form $\sum_{t'=t}^{T_n} r_{t'}^n - b$ is called the advantage of taking $a_t$ under $s_t$, which is parameterized by a neural network $A^\theta(s_t, a_t)$ in deep policy gradient algorithms. We assume an simulator is well-calibrated that the policy network $\pi_\theta$ trained in the simulated environment is trustful to initialize the parameters in the isomorphic real controller $\pi_{\theta'}$, and $\theta'$ is fixed for an episode (say, 1 hour). This 1 hour' real-world operation generates a SARSA trajectory $\tau^{\theta'}$, which can be used to approximate the more realistic advantage value function $A^{\theta'}(s, a)$. Usually, $A^{\theta'}(s, a)$ and $A^\theta(s, a)$ are different because of the "reality gap" between the state transition distributions of simulation and reality. As a consequence, we can just use the "real" trajectories collected from $\pi_\theta$ to slightly modify the policy network, and this process is awfully inefficient that it is synchronized with the actual clock.

Importance Sampling (Munos et al., 2016) provides a possibility of learning from simulated trajectories, whose core idea is formulated as,

$$E_{x \sim p}[f(x)] = \int f(x)p(x)dx = \int f(x)\frac{p(x)}{q(x)}q(x)dx = E_{x \sim q}[f(x)\frac{p(x)}{q(x)}] \qquad (6.3)$$

Thus the gradient $\nabla_{\theta'} J(\theta')$ of $J(\theta')$ can be rewritten as a weighted importance form as,

$$\nabla_{\theta'} J(\theta') = E_{\tau \sim p_\theta(\tau)}\left[\frac{p_{\theta'}(a_t, s_t)}{p_\theta(a_t, s_t)} A^{\theta'}(a_t, s_t) \nabla_\theta \log p_{\theta'}(a_t, s_t)\right] \qquad (6.4)$$

Furthermore, with the same detection technology and state definition, the probability of the occurrence of a specific state $s$ is approximately equal in two environments, therefore,

$$\frac{p_{\theta'}(a_t, s_t)}{p_\theta(a_t, s_t)} = \frac{p_{\theta'}(a_t | s_t) p_{\theta'}(s_t)}{p_\theta(a_t | s_t) p_\theta(s_t)} = \frac{p_{\theta'}(a_t | s_t)}{p_\theta(a_t | s_t)} \qquad (6.5)$$

Together, the expected return function of $\pi_{\theta'}$ is conditional on $\theta$,

$$J(\theta' | \theta) = E_{\tau \sim p_\theta(\tau)}\left[\frac{p_{\theta'}(a_t | s_t)}{p_\theta(a_t | s_t)} A^{\theta'}(a_t, s_t)\right] \qquad (6.6)$$

Theoretically, with a "real" SARSA trajectory, we can first estimate the advantage network of $A^{\theta'}$, then activate the simulator to generate samples $(a_t, s_t) \sim p_\theta(\tau)$ as many as possible, followed up with "mixed" parameter updating using Importance Sampling. Imagine two parallel worlds, in the real world, $\pi_{\theta'}$ collects samples via interacting with the physical environment, while in the virtual world, multiple simulators run simultaneously with much quicker speed, producing enough i.i.d trajectories to update $\theta'$ continuously. However, the Importance Sampling mechanism fails when the difference between $\theta$ and $\theta'$ is out of an acceptable range. When to stop the sampling process is stated in (Schulman et al., 2017).